\documentclass{article}

\usepackage{iclr2016_conference,times}
\usepackage{times}
\usepackage{epsfig}
\usepackage{epstopdf}
\usepackage{graphicx}
\usepackage{amsmath}
\usepackage{amssymb}
\usepackage{caption}
\usepackage{subcaption}
\usepackage{gensymb}
\usepackage{wrapfig}
\DeclareRobustCommand\onedot{\futurelet\@let@token\@onedot}
\def\@onedot{\ifx\@let@token.\else.\null\fi\xspace}

\iclrfinalcopy

\graphicspath{{figs/}{figs/KNN_RGBD_Wellposed_Train_Dist2/}{figs/KNNDist2_trainOnly_PASCAL3d/}} 
 
\usepackage[pagebackref=true,breaklinks=true,letterpaper=true,colorlinks,bookmarks=false]{hyperref}

\title{Digging Deep into the layers of CNNs: In Search of How CNNs Achieve View Invariance}

\author{    
	$\,\,\,\,\,\,\,\,\,\,\,\,\,\,$\textbf{Amr Bakry$^*$} \\
	{\tt\small amrbakry@cs.rutgers.edu}
	\and
	\textbf{Mohamed Elhoseiny$^*$}\\
	{\tt\small m.elhoseiny@cs.rutgers.edu}\\
	\and
	\textbf{Tarek El-Gaaly$^*$}\\ 
	{\tt\small tgaaly@cs.rutgers.edu}
	\and
	\textbf{Ahmed Elgammal}\\
	{\tt\small elgammal@cs.rutgers.edu}
	\and
	$\,\,\,\,\,\,\,\,\,\,\,\,\,\,\,\,\,\,\,\,\,\,\,\,\,\,\,\,\,\,\,\,\,\,\,\,\,\,\,\,\,\,\,\,\,\,\,\,\,\,\,\,\,\,\,\,\,\,\,\,$* indicates Co-first authors\\
	$\,\,\,\,\,\,\,\,\,\,\,\,\,\,\,\,\,\,\,\,\,\,\,\,\,\,\,\,\,\,\,\,\,\,\,\,\,\,\,\,\,\,\,\,\,\,\,\,\,\,\,\,\,\,\,\,\,\,\,\,$Computer Science Department, Rutgers University\\
}

\begin{document}

\maketitle

\begin{abstract}
This paper is focused on studying the view-manifold structure in the feature spaces implied by the different layers of Convolutional Neural Networks (CNN). There are several questions that this paper aims to answer: 
Does the learned CNN representation achieve viewpoint invariance? How does it achieve viewpoint invariance? Is it achieved by collapsing the view manifolds, or separating them while preserving them? At which layer is view invariance achieved? How can the structure of the view manifold at each layer of a deep convolutional neural network be  quantified experimentally? How does fine-tuning of a pre-trained CNN on a multi-view dataset affect the representation at each layer of the network? In order to answer these questions we propose a methodology to quantify the deformation and degeneracy of view manifolds in CNN layers. We apply this methodology and report interesting results in this paper that answer the aforementioned questions.

\end{abstract}
\section{Introduction}
\label{S:Intro}


Impressive results have been achieved recently with the application of Convolutional Neural Networks (CNNs) in the tasks of object categorizations ~\citep{krizhevsky2012imagenet} and detection~\citep{Sermanet_EZMFL_2013, Malik_rcnn_2013}. Several studies recently investigated different properties of the learned representations at different layers of the network, {\emph{e.g.}} ~\citep{Yosinski_howtransferable_2014,fergus_visualizecnn_2013,zisserman_devilindetails_2014}. One fundamental question is how CNN models achieve different invariances. It is well understood that consecutive convolution and pooling layers can achieve translation invariant.   Training CNN networks with a large dataset of images, with arbitrary viewpoints and arbitrary illumination, while optimizing the categorization loss  helps to achieve viewpoint invariant and illumination invariant. 

In this paper we focus on studying the viewpoint invariant properties of CNNs. In many applications, it is desired to estimate the pose of the object, for example for robot manipulation and scene understanding. Estimating pose and object categorization are tasks that contradict each other; estimating pose requires a representation capable of capturing the viewpoint variance, while viewpoint invariance is desired for categorization. Ultimately, the vision system should achieve a representation that can factor out the viewpoint for categorization and preserve viewpoint for pose estimation.

\begin{wrapfigure}{R}{0.7\textwidth}
	  \vspace{-4mm}
  \centering
  \includegraphics[height=2in]{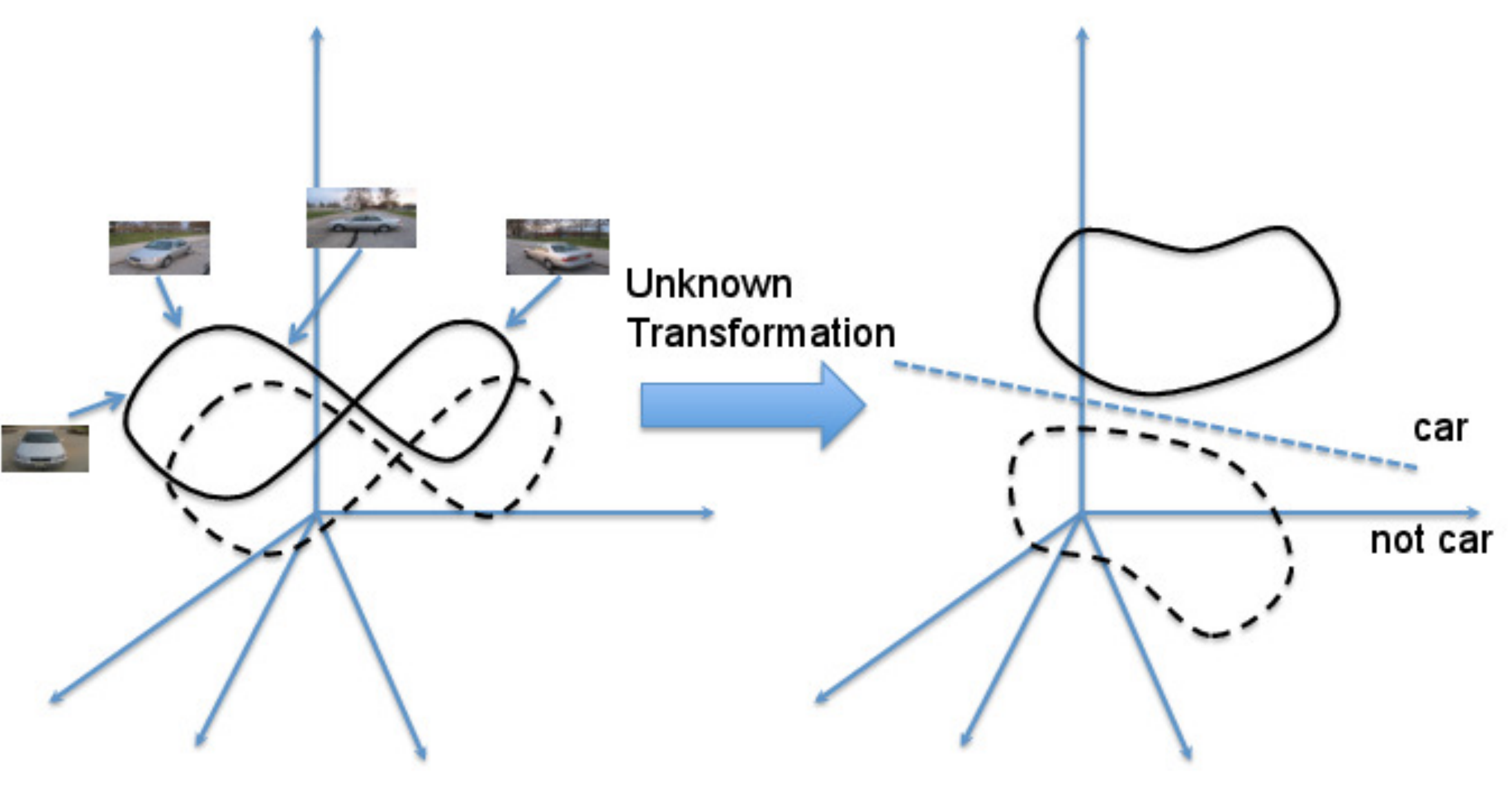}
    \vspace{-4mm}
  \caption{Illustration of DiCarlo and Cox model~\citep{dicarlo2007untangling}: Left: tangled manifolds of different objects in early vision areas. Right: untangled (flattened) manifold representation in IT}
  \vspace{-4mm}
  \label{DiCarloModel}
\end{wrapfigure}

The biological vision system is able to recognize and categorize objects under wide variability in visual stimuli, and at the same time is able to recognize object pose. It is clear that images of the same object under different variability, in particular different views, lie on a low-dimensional manifold in the high-dimensional visual space defined by the retinal array ($\sim$100 million photoreceptors and $\sim$1 million retinal ganglion cells). \cite{dicarlo2007untangling} hypothesized that the ability of our brain to recognize objects, invariant to different viewing conditions, such as viewpoint, and at the same time estimate the pose, is fundamentally based on untangling the visual manifold encoded in neural population in the early vision areas (retinal ganglion cells, LGN, V1). They suggested that this is achieved through a series of successive transformation (re-representation) along the ventral stream (V1,V2, V4, to IT) that leads to an untangled population at IT. Despite this, it is unknown how the ventral stream achieves this untangling. They argued that since IT population supports tasks other than recognition, such as pose estimation, the manifold representation is some how \textit{'flattened'} and \textit{'untangled'} in the IT layer. DiCarlo and Cox's hypothesis is illustrated in Figure~\ref{DiCarloModel}. They stress that the feedforward cascade of neural re-representation is a way to untangle the visual manifold. 

%

Inspired by recent impressive results of CNNs and by DiCarlo and Cox's hypothesis ~\citep{dicarlo2007untangling} on manifold untangling, this paper focuses on studying the view-manifold structure in the feature spaces implied by the different layers of CNNs. There are several questions that this paper aims to answer: 
\textbf{1.} Does the learned CNN representations achieve viewpoint invariance? If so, how does it achieve viewpoint invariance? Is it by collapsing the view manifolds, or separating them while preserving them? At which layer is the view invariance achieved?
\textbf{2.} How to experimentally quantify the structure of the viewpoint manifold at each layer of a deep convolutional neural network?
\textbf{3.} How does fine-tuning of a pre-trained CNN, optimized for categorization, on a multi-view dataset, affect the representation at each layer of the network?

In order to answer these questions, we present a methodology that helps to get an insight about the structure of the viewpoint manifold of different objects as well as the combined object-view manifold in the layers of CNN. 
We conducted a series of experiments to quantify the ability of different layers of a CNN to either preserve the view-manifold structure of data or achieve a view-invariant representation. 

The contributions of the paper are as follows: 
(1) We propose a methodology to quantify and get insight into the manifold structures in the learned representation at different layers of CNNs.
(2) We use this methodology to analyze the viewpoint manifold of pre-trained CNNs.
(3) We study the effect of transfer learning a pre-trained network with two different objectives (optimizing category loss vs. optimizing pose loss) on the representation.
(4) We draw important conclusions about the structure of the object-viewpoint manifold and how it coincides with DiCarlo and Cox's hypothesis.

The paper begins by reviewing closely related works. Section \ref{S:ProbDef} defines the problem, experimental setup, and the basic CNN network that our experiments are based upon. Section \ref{S:ManifoldAnalysis} introduces our methodology of analysis. Sections~\ref{S:Analysis} and~\ref{S:FTAnalysis} describe the findings on the pre-trained network and the fine-tuned networks respectively. The conclusion section summarizes our findings.

\section{Related Work}
\label{S:RelWork}


LeCun {\emph{et al.}} has widely used CNNs for various vision tasks \citep{Sermanet_EZMFL_2013, kavukcuoglu_SBGML_2010, Jarrett_whatisbest_09,Ranzato_unsupervised_2007, LeCun_learningmethods_2004}. The success of CNNs can be partially attributed to these efforts, in addition to training techniques that have been adopted. \cite{krizhevsky2012imagenet} used a CNN in the ImageNet Challenge 2012 and achieved state-of-the-art accuracy. Since then, there have been many variations in CNN  architectures and learning techniques within different application contexts. In this section we mainly emphasize related works that focused on bringing an understanding of the representation learned at the different layers of CNNs and related architectures.

\cite{Yosinski_howtransferable_2014} studied how CNN layers transition from general to specific. An important finding in this study is that learning can be transferred, and by using fine-tuning, performance is boosted on novel data. Other transfer learning examples include~\citep{razavian2014cnn,donahue2013decaf,agrawal2014analyzing}.   \cite{fergus_visualizecnn_2013} investigated the properties of CNN layers for the purpose of capturing object information. This study is built on the premise that there is no coherent understanding of why CNNs work well or how we can improve them. Interesting visualizations were used to explore the functions of layers and the intrinsics of categorization. The study stated that CNN output layers are invariant to translation and scale but not to rotations. The study in \citep{zisserman_devilindetails_2014} evaluated different deep architectures and compared between them. The effect of the output-layer dimensionality was explored.

\section{Problem Definition and Experimental Setup}
\label{S:ProbDef}


It is expected that multiple views of an object lie on intrinsically low-dimensional manifolds ({\emph view manifold}\footnote{we use the terms \textit{view manifold }and \textit{viewpoint manifold }interchangeably}) in the input space. View manifolds of different instances and different objects are spread out in this input space, and therefore form jointly what we call the {\emph object-view manifold}. The input space here denotes the $R^{N\times M}$ space induced by an input image of size $N \times M$, which is analogous to the retinal array in the biological system. For the case of a viewing circle(s), the view manifold of each object instance is expected to be a 1-dimensional closed curve in the input space. The recovery of the category and pose of a test image reduces to finding which of the manifolds this image belongs to, and what is the intrinsic coordinate of that image within that manifold. This view of the problem is shared among manifold-based approaches such as~\citep{murase95visual,zhang_aaai_2013,bakry_untangling_2014}

The ability of a vision system to recover the viewpoint is directly related to how the learned representation preserves the view manifold structure. If the transformation applied to the input space yields a representation that  results in collapsing the view manifold, the system will no longer be able to discriminate between different views. Since each layer of a deep NN re-represents the input in a new feature space, the question would be how the re-representations deform a manifold that already exists in the input space.  A deep NN would satisfy the hypothesis of \textit{'flattening'} and \textit{'untangling'} by ~\cite{dicarlo2007untangling}, if the representation in a given layer separates the view manifolds of different instances, without collapsing them, in a way to be able to put a separating hyperplanes between different categories. 
Typically CNN layers exhibit general-to-specific feature encoding, from Gabor-like features and color blobs at low layers to category-specific features at higher layers~\citep{fergus_visualizecnn_2013}. We can hypothesize that for the purpose of pose estimation, lower layers should hold more useful representations that might preserve the view manifold and be better for pose estimation. But which of these layers would be more useful, and where does the view-manifold collapse to view-invariance.
\begin{wrapfigure}{R}{0.6\textwidth}
\vspace{-7mm}
\begin{minipage}{\hsize}
  \includegraphics[width=\linewidth]{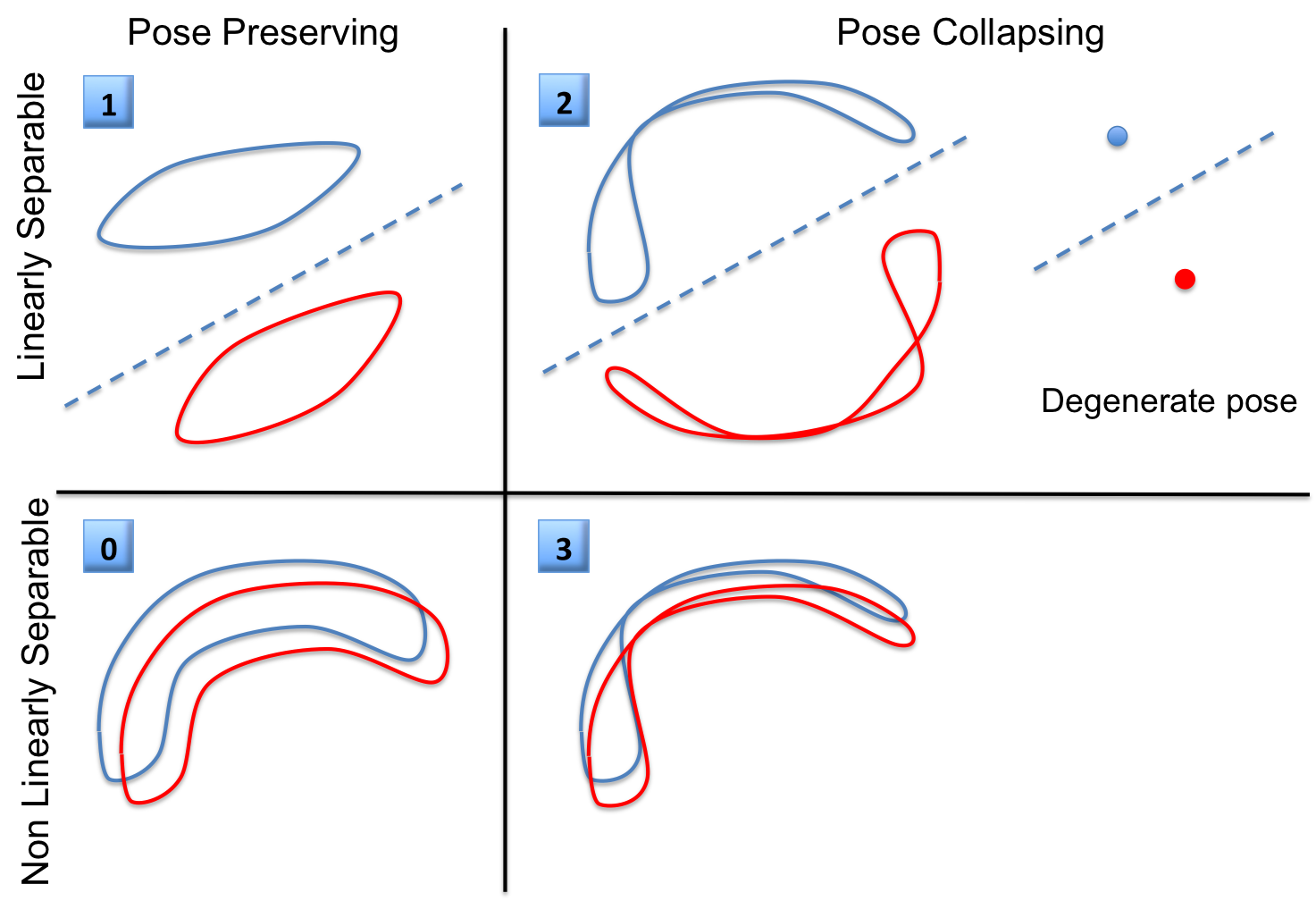}
  \caption{\footnotesize Sketches of four hypotheses about possible structures of the view manifolds of two objects in a given feature space. 
  	}
  \label{F:Four_hypotheses}
\end{minipage}
\vspace{-4mm}
\end{wrapfigure}

There are different hypotheses we can make about how the view manifolds of different objects are arranged in the feature space of a given layer. These hypotheses are shown in Figure~\ref{F:Four_hypotheses}. We arrange these hypotheses based on linear separability of the different objects' view manifolds and the preservation of the view manifolds. Case 0 is the non-degenerate case where the visual manifolds preserve the pose information but are tangled and there is no linear separation between them (this might resemble the input space, similar to left case in Figure~\ref{DiCarloModel}). Case 1 is the ultimate case where the view manifolds of different objects are preserved by the transformation and are separable (similar to the right case in Figure~\ref{DiCarloModel}).  Case 2 is where the transformation in the network leads to separation of the object's view manifold at the expense of collapsing these manifolds to achieve view invariance. Collapsing of the manifolds can be to different degrees, to the point where each object's view manifold can be mapped to a single point. Case 3 is where the transformation results in more tangled manifolds (pose collapsing and non-separable). It is worth to notice that both cases 1 and 2 are view invariant representations. However, it is obvious that case 1 would be preferred since it also facilitates pose recovery. It is not obvious whether optimizing a network with a categorization loss result in case 1 or case 2. 
Getting an insight about which of these hypotheses are true in a given layer of a CNN is the goal of this paper. In Section~\ref{S:ManifoldAnalysis} we propose a methodology to get us to that insight.%

\begin{figure*}[!tb]
	\centering
\begin{minipage}{0.9\hsize}
	\centering
	\includegraphics[width=\linewidth,height=0.55\linewidth]{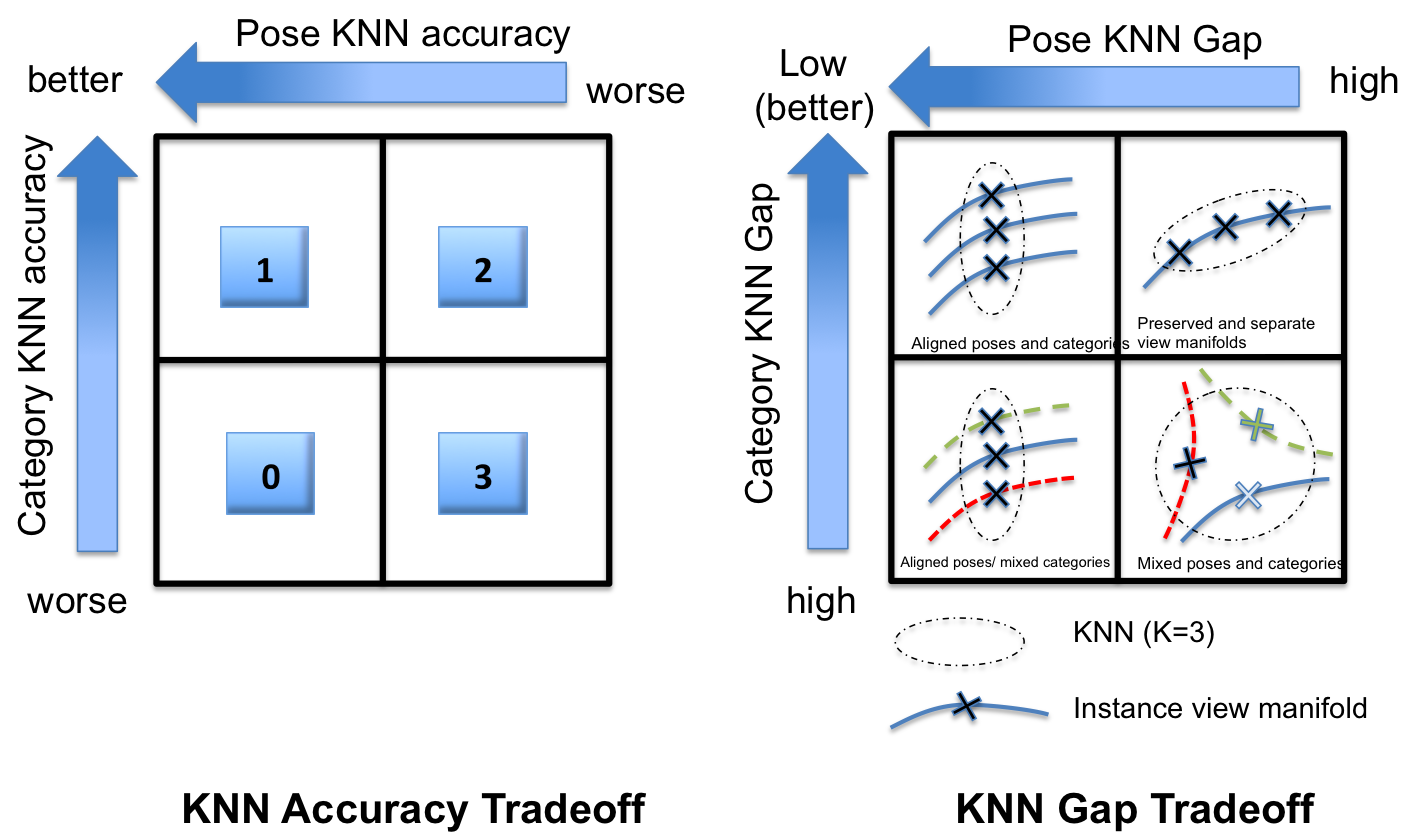}

\end{minipage}
	\vspace{-3mm}
	\caption{\footnotesize KNN Tradeoffs: accuracy tradeoff between category and pose estimation using KNN. This cartoon illustrates the global measurements, see Section~\ref{sec:global_measurements} for full details.}
	\label{F:KNN}
\end{figure*}

\subsection{Experimental Settings}
To get an insight into the representations of the different layers and answer the questions posed in Section~\ref{S:Intro} we experiment on two datasets: I) RGB-D dataset~\citep{lai2011large}, II) Pascal3D+ dataset~\citep{xiang_pascal3d_2014}. We selected the RGB-D dataset since it is the largest available multiview dataset with the most dense viewpoint sampling. The dataset contains 300 instances of tabletop objects (51 categories). Objects are set on a turntable and captured by an Xbox Kinect sensor (Kinect 2010) at 3 heights (30$^\circ$, 45$^\circ$ and 60$^\circ$ elevation angles). The dense view sampling along each height is essential for our study to guarantee good sampling of the view manifold. We ignore the depth channel and only used the RGB channels. 

Pascal3D+ is very challenging because it consists of images ``in the wild'', in other words, images of object categories exhibiting high variability, captured in uncontrolled settings and under many different poses. Pascal3D+ contains 12 categories of rigid objects selected from the PASCAL VOC 2012 dataset \citep{Pascal}. These objects are annotated with 3D pose information ({\emph{i.e}}, azimuth, elevation and distance to camera). Pascal3D+ also adds 3D annotated images of these 12 categories from the ImageNet dataset \citep{imNet09}. The {\it bottle} category is omitted in state-of-the-art results. This leaves 11 categories to experiment with. There are about 11,500 and 7,000 training images in ImageNet and Pascal3D+ subsets, respectively.  For testing, there are about 11,200 and 6,900 testing images for ImageNet and Pascal3D+, respectively. On average there are about 3,000 object instances per category in Pascal3D+, making it a challenging dataset for estimating object pose.

The two datasets provide different aspect of the analysis. While the RGB-D provides dense sampling of each instance's view manifold,  Pascal3D+ dataset contains only very sparse sampling.  Each instance is typically imaged from a single viewpoint, with multiple instances of the same category sampling the view manifold at arbitrary points.
 Therefore, in our analysis we use the RGB-D dataset to analyze each instance viewpoint manifold and the combined object-viewpoint manifolds, while the Pascal3D provides analysis of the viewpoint manifold at the category level.

\noindent{\bf Evaluation Split:}
For our study, we need to make sure that the objects we are dealing with have non-degenerate view manifolds. We observed that many of the objects in the RGB-D dataset are ill-posed, in the sense that the poses of the object are not distinct. This happens when the objects have no discriminating texture or shape to be able to identify the different poses (\emph{e.g.} a texture-less ball, apple or orange on a turntable). This will cause view manifold degeneracy. Therefore we select 34 out of the 51 categories as objects that possess pose variations across the viewpoints, and thus are not ill-posed with respect to pose estimation.

We split the data into training, validation and testing. Since in this datasets, most categories have few instances, we left out two random object instances per category, one for validation and one for testing. In the case where a category has less than 5 instances, we form the validation set for that category by randomly sampling from the training set. Besides the instance split, we also left out all the middle height for testing. Therefore, the testing set is composed of unseen instances and unseen heights and this allows us to more accurately  evaluate the capability of the CNN architectures in discriminating categories and estimating pose of tabletop objects. 
\subsection{Base Network: Model0}
The base network we use is the Convolutional Neural Network described in \cite{krizhevsky2012imagenet} and winner of LSVRC-2012 ImageNet challenge~\citep{li_imagenet_2014}. The CNN was composed of 8 layers (including 1000 neuron output layer corresponding to 1000 classes). We call these layers in order: Conv1, Pool1, Conv2, Pool2, Conv3, Conv4, Conv5, Pool5, FC6, FC7, FC8 where Pool indicates Max-Pooling layers, Conv indicates layers performing convolution on the previous layer and FC indicates fully connected layer. The last fully connected layer (FC8) is fed to a 1000-way softmax, which produces a distribution over the category labels of the dataset. 

\section{Methodology}
\label{S:ManifoldAnalysis}


The  goal of our methodology is two-folds: (1) study the transformation that happens to the viewpoint manifold of a specific object instance at different layers, (2) study the structure of the combined object-view manifold at each layer to get an insight about how tangled or untangled the different objects' viewpoint manifolds are. Both these approaches will get us an insight to which of the hypotheses explained in Section~\ref{S:ProbDef} is correct at each layer, at least relatively by comparing layers.
This section introduces our methodology, which consists of two sets of measurements to address the aforementioned two points.  First, we introduce instance-specific measurements that quantify the viewpoint manifold in the different layers to help understand whether the layers preserve the manifold structure. 
{\emph We performed extensive analysis on synthetic manifold data to validate the measures, see Appendix~\ref{sec:synthetic_dataset}.}
Second, we introduce empirical measurements that are designed to draw conclusions about the global object-viewpoint manifold (involving all instances).

\subsection{Instance-Specific View Manifold Measurements}



Let us denote the input data (images taken from a viewing circle and their pose labels) for a specific object instance  as $\{(x_i \in \mathbb{R}^D , \theta_i \in [0,2 \pi] ) , i=1 \cdots N \}$, where $D$ denotes the dimensionality of the input image to the network, and $N$ is the number of the images, which are equally spaced around the viewing circle. These images form the view manifold of that object in the input space denoted by $\mathcal{M}=\{x_i\}_1^N$. Applying each image to the network will result in a series of nonlinear transformations. Let us denote the transformation from the input to layer $l$ by the function $f_l(x):  \mathbb{R}^D \rightarrow {R}^{d_l} $ where $d_l$ is the dimensionality of the feature space of layer $l$.  With an abuse of notation we also denote the transformation that happens to the manifold $\mathcal{M}$ at layer $l$ by $\mathcal{M}^l=f_l(\mathcal{M})= \{ f_l(x_i)\}_1^N $. 
After centering the data by subtracting the mean, let $\mathbf{A}^l=[\acute{f}_l(x_i) \cdots \acute{f}_l(x_N)]$ be the centered feature matrix at layer $l$ of dimension $d_l \times N$, which corresponds to the centered transformed images of the given object. We call $\mathbf{A}^l$ the sample matrix in layer $l$.


Since the dimensionality $d_l$ of the feature space of each layer varies, we need to factor out the effect of the dimensionality. Since $N \ll d_l$ the transformed images on all the layers lie on  subspaces of dimension $N$ in each of the feature spaces. Therefore, we can change the bases to describe the samples using $N$ dimensional subspace, {\emph{i.e}}, we define the $N\times N$ matrices $\mathbf{\hat{A}}^l = \mathbf{U}^\textsf{T} \mathbf{A} $ where  $\mathbf{U} \in \mathbb{R}^{d_l \times N}$ are the orthonormal bases spanning the column space of $\mathbf{A}^l$ (which we can get by SVD of $\mathbf{A}^l=  \mathbf{U} \mathbf{S} \mathbf{V}^\textsf{T} $). This projection rotates the samples at each layer without changing the manifold geometric or neighborhood properties. Then the following measures will be applied to the $N$ transformed images, representing the view manifold of each object instance individually. To obtain an overall measures for each layer we will average these measures over all the object instances. 


\smallskip
\noindent{\emph 1) Measure of spread - Nuclear Norm:} 
There are several possible measures of the spread of the data in the sample matrix of each view manifold. We use the nuclear norm (also known as the trace norm~\citep{HornMatrixAnalysisBook}) defined as $|| \mathbf{A} ||_* = Tr (\sqrt{\mathbf{A}^\textsf{T}\mathbf{A}}) = \sum_{i=1}^N \sigma_i$, {\emph{i.e}}, it measures the sum of the singular values of  $\mathbf{A}$. 

\smallskip
\noindent {\emph 2) Subspace dimensionality measure - Effective-p:} counts the effective dimensionality of the subspace where the view manifold lives. Smaller number means that the view manifold lives in lower dimensional subspace. We define {\emph Effective-p} as the minimum number of singular values (in decreasing order) that sum up to more that or equal to $p\%$ of the nuclear norm, i.e, $\text{Effective}-p = \sup \{n: {\sum_{i=1}^n \sigma_i}/{\sum_{i=1}^N \sigma_i} \leq p/100 \}$.

\smallskip  
\noindent{\emph 3) Alignment Measure -  KTA:}
Ideally the view manifold resulting of the view sitting of the studied datasets is a single-dimensional closed curve in the feature space, which can be thought as a deformed circle~\citep{zhang_aaai_2013}. This manifold can be degenerate in the ultimate case to a single point in case of a texture-less object.  The goal of this measurement is to quantify how the transformed manifold locally preserves the original manifold structure. To this end we compare the kernel matrix of the transformed manifold at layer $l$, denote by $\mathbf{K}^l_n$, with the kernel matrix of the an embedding of the ideal view manifold on unit circle, denote by $\mathbf{K}_n^\circ$, where $n$ indicates the local neighborhood size used in constructing the kernel matrix. We construct the neighborhood based on pose labels. 

Given these two kernel matrices we can define several convergence measures.  We use {\emph Kernel Target Alignment (KTA)} which has been used in the literature for kernel learning~\citep{Kandola_spectralkernel_2001}. It finds a scale invariant dependency between two normalized kernel matrices\footnote{We also experimented with HSIC~\citep{gretton2005measuring}, however HSIC is not scale invariant and not designed to compare data in different feature spaces. Therefore, HSIC did not give any discriminative signal}. Therefore, we define the alignment of the transformed view manifold $\mathcal{M}^l$ at layer $l$  with the ideal manifold as $KTA_n(\mathcal{M}^l) =  {<\mathbf{K}_n^l,\mathbf{K}_n^\circ>_F}/{(|| \mathbf{K}_n^l||_F  || \mathbf{K}_n^\circ ||_F )}$.

\smallskip
\noindent{\emph 4) KPLS-regression measures:}  Kernel Partial Least Squares (KPLS) \citep{Rosipal_kpls_2002} is a supervised regression method. KPLS iteratively extracts the set of principal components of the input kernel that are most correlated with the output 
. We use KPLS to learn mapping $\mathbf{K}^l_n \rightarrow \mathbf{K}_n^\circ$ from the transformed view manifold kernel (input kernel) to the unit circle kernel (output kernel). We enforce this mapping to use maximum of $d\ll N$ principal components (we used $d=5$). 
Then we define {\emph KPLS-Regression Error}, which uses the Normalized Cross Correlation to quantify the mapping correctness.



\smallskip
\noindent{\emph 5) TPS-linearity measure:} In this measure we learn a regularized Thin Plate Spline (TPS) non-linear mapping~\citep{duchon1977splines} between the unit circle manifold and each $\mathcal{M}^l$. The reason for using TPS in particular is that the mapping has two parts: affine (linear polynomial) and nonlinear part. 
Analysis of the two parts will tell us if the mapping is mostly linear or nonlinear. We use the {\emph reciprocal-condition number (rcond)} of the sub coefficient matrices corresponding to the affine and the nonlinear part as a measure of the linearity of the transformation. \footnote{More details and definitions about KPLS and TPS based measurements in Appendix~\ref{sec:moreMeaurements}.}

\subsection{Global Object-Viewpoint Manifold Measures}
\label{sec:global_measurements}
To achieve an insight about the global arrangement of the different objects' view-manifolds in a given feature (layer) space, we use the following three empirical measurements:

 \smallskip
\noindent{\emph 6) Local Neighborhood Analysis:}
To evaluate the local manifold structure we also evaluate the performance of nearest neighbor classifiers for both category and pose estimation, with varying size of the neighborhood. This directly tell us whether the neighbors of a given point are from the same category and/or of similar poses. KNN for categorization cannot tell us about the linear separability of classes. However evaluating the pose estimation in neighborhood of a datapoint gives us an insight about how the view manifolds are preserved, and even whether the view manifolds of different instances are aligned. To achieve this insight we use two different measurements:
\textbf{KNN-Accuracy}: the accuracy of KNN classifiers for  category and pose estimation. 
\textbf{KNN-Gap}: the drop in performance of each KNN classifier as the neighborhood size increases. In our experiments we increase K from 1 to 9. Positive gap indicates a drop (expected) and negative gap indicates improvement in performance.

The interaction between these two measures and how they tell us about the manifold structure is illustrated in Fig~\ref{F:KNN}. The contrast between the accuracy of the KNN classifiers for pose and category directly implies which of the hypotheses in Figure~\ref{F:Four_hypotheses} is likely. The analysis of KNN-Gap (assuming good 1-NN accuracy) gives further valuable information. As the KNN-gap reaches zero in both category and pose KNN classifiers, this implies that neighborhoods are from the same category and has the same pose, which indicates that the representation aligns the view manifolds of different  instances of the same category.  If the view manifolds of such instances are preserved and separated in the space, and the neighbors of a given point are from the same instance, this would imply small gap in the category KNN classifier and bigger gap in pose KNN classifier.  Low gap in pose KNN vs high gap in category CNN implies the representation aligns view manifolds of instances of different categories. A high gap in both obviously implies the representation is tangling the manifolds such that a small neighborhood contains points from different categories and different poses. Notice that this implications are only valid when the 1-NN accuracy is high.

\smallskip
\noindent{\emph 7) L-SVM:}
For a test image $x$ transformed to the $l$-th layer's feature space, $f_l(x)$, we compute the performance of a linear SVM classifier trained for categorization. Better performance of such a classifier directly implies more linear separability between different view manifolds of different categories.

\smallskip
\noindent{\emph 8) Kernel Pose Regression:}
To evaluate whether the pose information is preserved in a local neighborhood of a point in a given feature space we evaluate the performance of kernel ridge regression for the task of pose estimation. Better performance implies better pose-preserving transformation, while poor performance  indicates pose-collapsing transformation. The combination of L-SVM and kernel regression should be an indication to which of the hypotheses in Figure~\ref{F:Four_hypotheses} is likely to be true.


\section{Analysis of the Pre-trained Network}
\label{S:Analysis}
\vspace{-5pt}
\subsection{Instance View Manifold Analysis}

\begin{figure*}[h] 
\centering
\begin{subfigure}[b]{0.33\hsize}
  \caption{\scriptsize Nuclear Norm}
    \vspace{-0.25em}
  \centering
    \includegraphics[width=\textwidth,height=.6\textwidth]{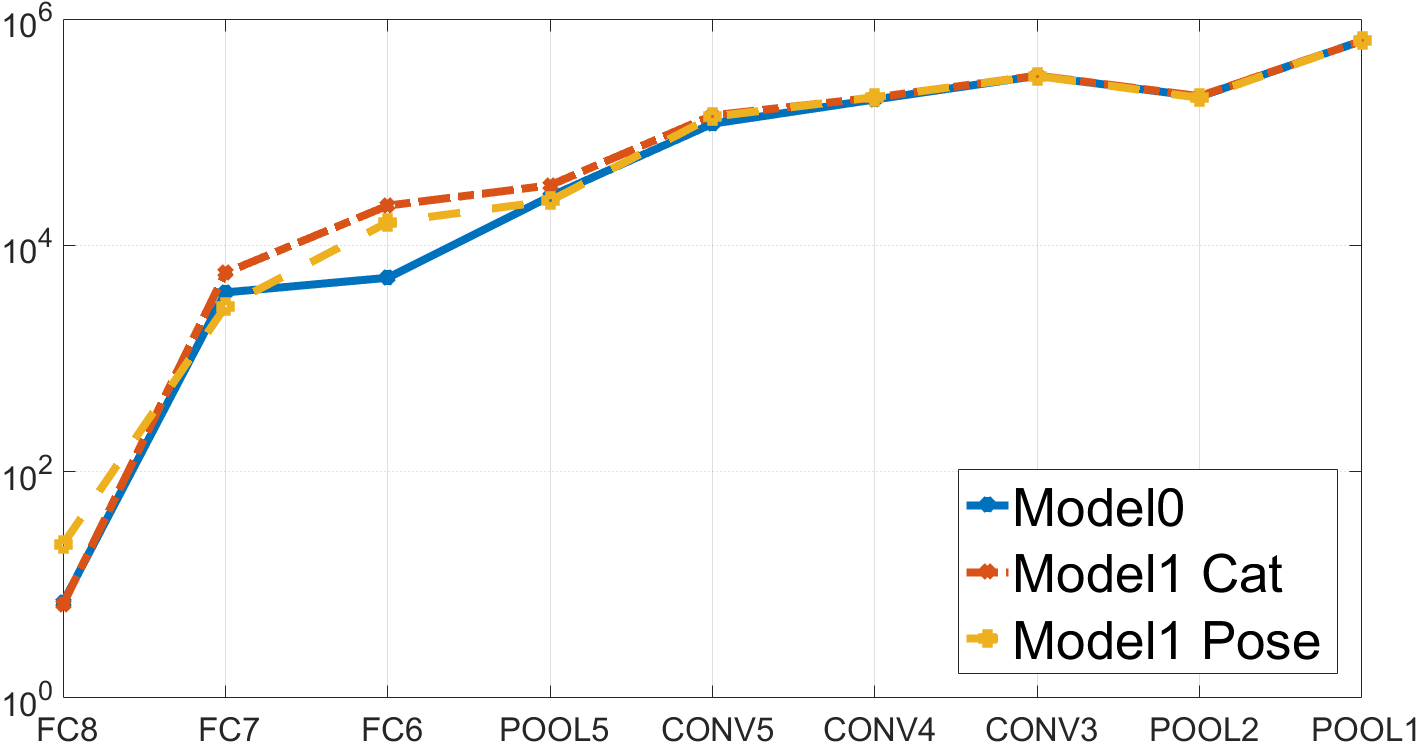}
        \label{SV_sum1}
\end{subfigure} 
\hspace{-0.5em}
\begin{subfigure}[b]{.33\hsize}
  \caption{\scriptsize Effective $90\%$ SV's}
    \vspace{-0.25em}
  \centering
    \includegraphics[width=\textwidth,height=.6\textwidth]{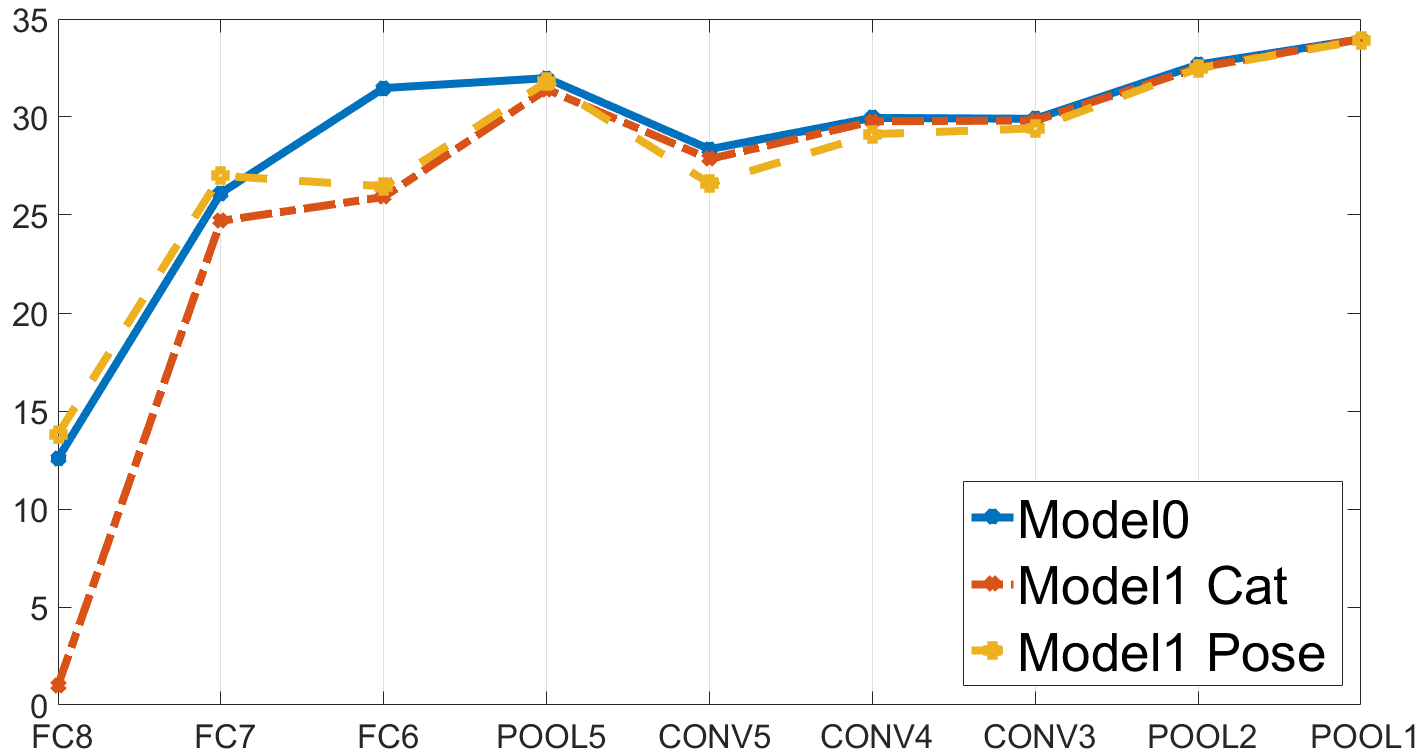}
        \label{EffectiveSV901}
\end{subfigure}
\hspace{-0.5em} 
\begin{subfigure}[b]{0.33\hsize}
  \caption{\scriptsize KTA}
    \vspace{-0.25em}
  \centering
    \includegraphics[width=\textwidth,height=.6\textwidth]{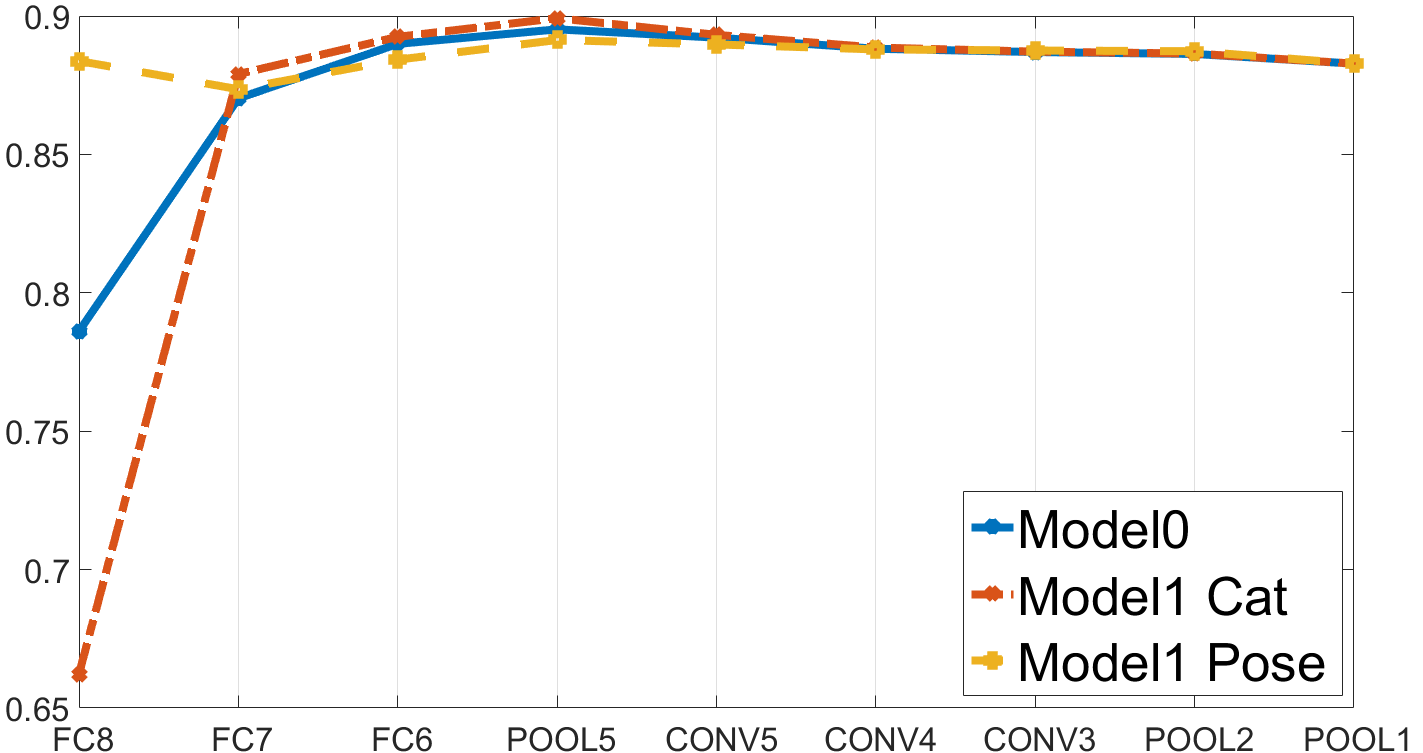}
    \label{kta1}
\end{subfigure}
\vspace{-1.5em}
\begin{subfigure}[b]{0.33\hsize}
  \caption{\scriptsize KPLS-Reg err}
    \vspace{-0.25em}
  \centering
    \includegraphics[width=\textwidth,height=.6\textwidth]{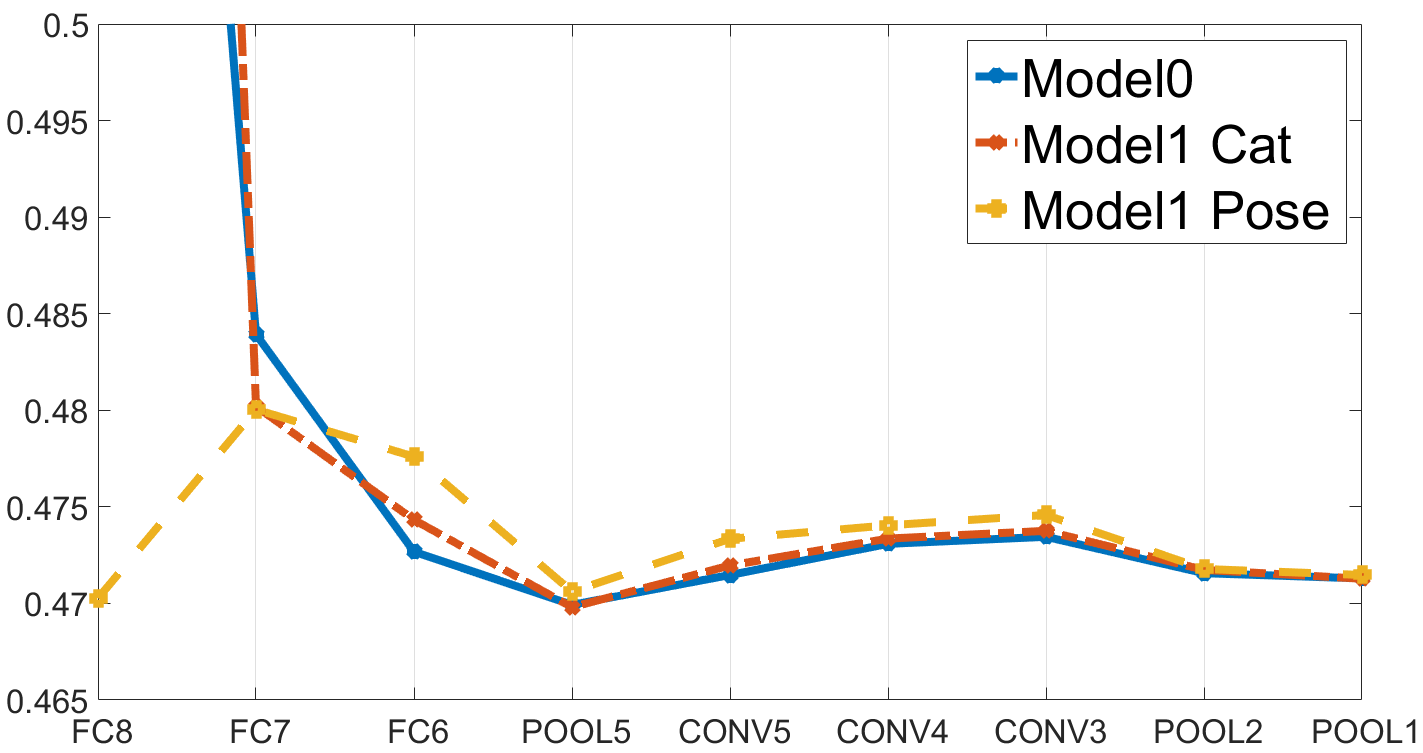}
    \label{regression_err1}
\end{subfigure}
\hspace{-0.5em}
\begin{subfigure}[b]{0.33\hsize}
  \caption{\scriptsize TPS-$RCond(poly)$}
  \vspace{-0.25em}
  \centering
    \includegraphics[width=\textwidth,height=.6\textwidth]{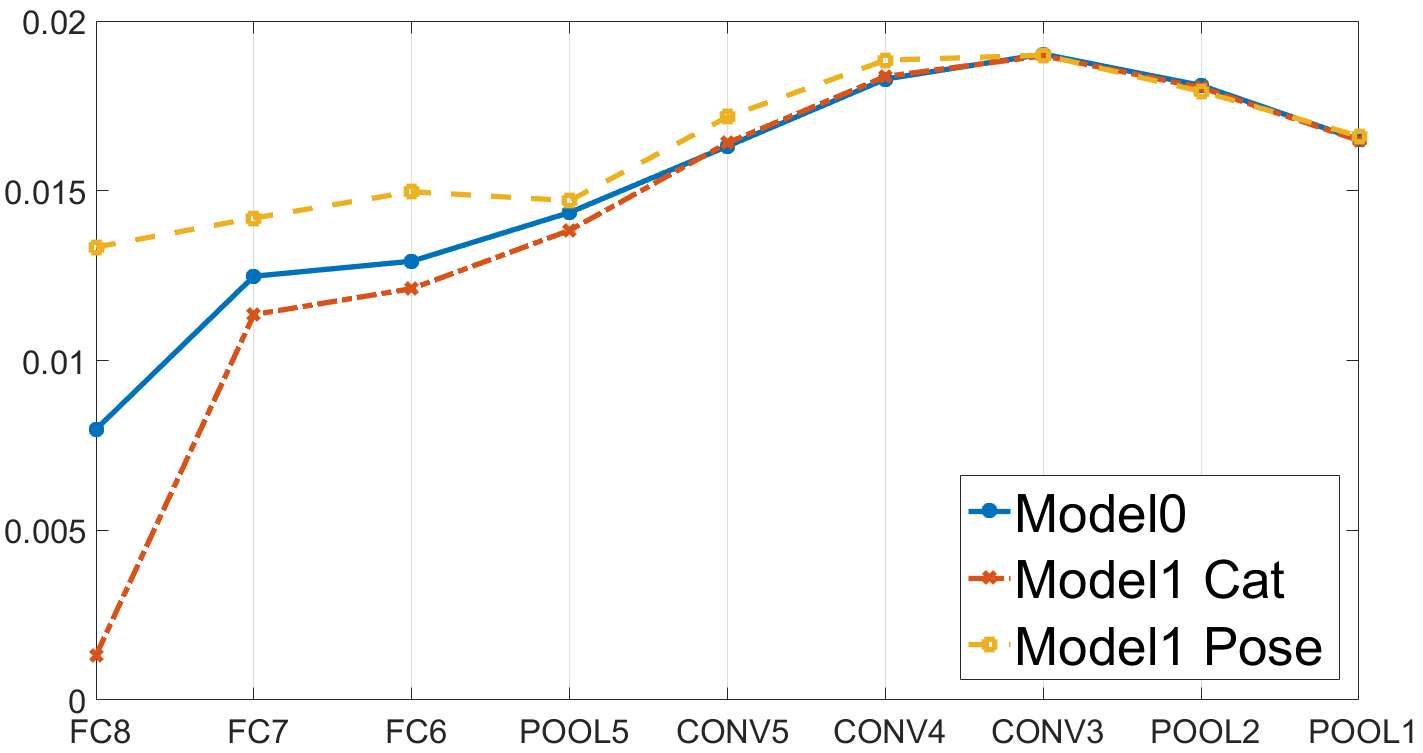}
    \label{tps_poly1}
\end{subfigure}
\vspace{-1mm}
\caption{\footnotesize RGB-D: Local Measurement analysis for the view-manifold. 
Every figure shows single measurement for three models (Model0, Model1Cat and Model1Pose) at different layers. 
	}
\label{F:AnalysisPlots2}
\end{figure*}

Figure \ref{F:AnalysisPlots2} shows the application of the instance-specific view manifold measurements on the images of the RGBD dataset when applied to a pre-trained network (Model0  - no fine-tuning).  This gives us an insight on the transformation that happens to the view manifold of each object instance at each layer of the network. Figure~\ref{SV_sum1} shows that the nuclear norm of the transformed view manifolds in Model0 is almost monotonically decreasing as we go higher in the network, which indicates that the view manifolds is more spread in the lower layers. In fact at the output layer of Model0 the nuclear norm becomes too small, which indicates that the view manifold  is collapsing to reach view invariant representation at this layer. Figure ~\ref{EffectiveSV901} ($p=90\%)$ shows that subspace dimension varies within a small range in the lower layers and it reduces dramatically in fully connected layers, which indicates that the network tries to achieve view invariance. The minimum is achieved at FC8 (even without fine tuning).   Figure~\ref{kta1} shows the KTA applied to Model0, where we can notice that the alignment is almost similar across the lower layers, with Pool5 having the maximum alignment, and then starts to drop at the very high layers. which indicates that after Pool5, the FC layers try to achieve view invariant. Fig~\ref{regression_err1} shows that KPLS regression error on Model0 dramatically reduces from FC8 down to Pool5, where Pool5 has the least error. In general the lower layers have less error. This indicates that the lower layers preserve higher correlation with the ideal manifold structure. Fig~\ref{tps_poly1} shows that the mapping is highly linear, which is expected because of the high dimensionality of the feature spaces. From Fig~\ref{tps_poly1} we can clearly notice that the lower layers has more better-conditioned linear mapping (plots for the nonlinear part is in Appendix~\ref{sec:moreMeaurements}.)

From these measurements we can conclude:
(1) The lower layers preserve the view manifolds. The manifolds start to collapse in the FC layers to achieve view invariance. Preserving the view manifold at the lower layers  is intuitive because of the nature of the convolutional layers. (2) The manifold at Pool5 achieves the best alignment with the pose labels. This is a less intuitive result; why does the representation after successive convolutions and pooling improves the view manifold alignment? even without seeing any dense view manifold in training, and even without any pose labels being involved in the loss.   The hypotheses we have to justify that Pool5 has better alignment than the lower layers is that Pool5 has better  translation invariant properties, which results in improvement of the view manifold alignment. 


\subsection{Global Object-View Manifold Analysis}
\begin{wrapfigure}{R}{0.7\textwidth}
\vspace{-7mm}
\centering	

\includegraphics[width=\hsize,height=.35\hsize]{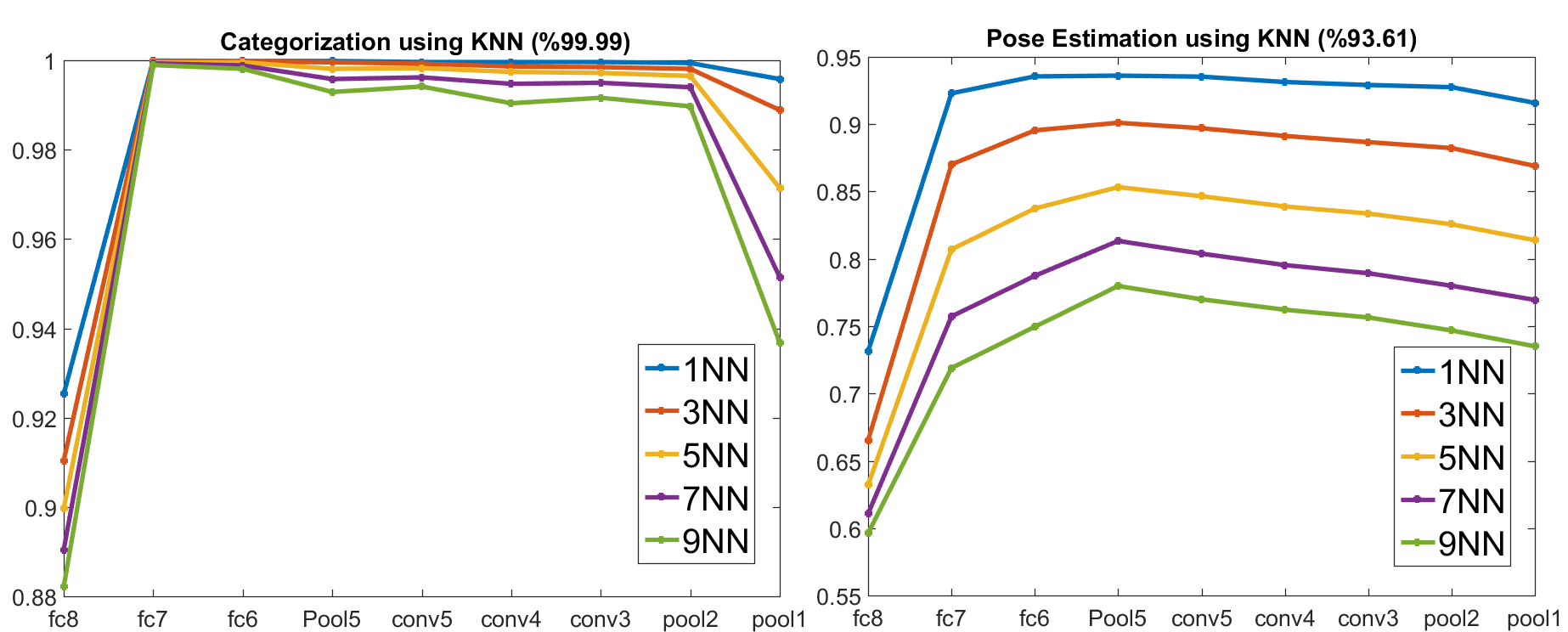}
\vspace{-4mm}
	\caption{\footnotesize RGB-D: KNN for categorization and pose estimation over the layers of pre-trained model (Model0). For $K = \{1,3,5,7,9\}$}.
\vspace{-4mm}
	\label{KNN_train_RGBD}
\end{wrapfigure}

To study view-manifold in the network layers, Figure~\ref{KNN_train_RGBD} shows the KNN accuracy for pose and category within training split, no test is used in this experiment. The category gap is reducing as we go up in the network up to FC7 (almost 0 gap at FC6 and FC7).  In contrast the gap is large at all layers for pose estimation. This indicates separation of the  instances' view manifolds where the individual manifolds are not collapsed (This is why as we increase the neighborhood, the category performance stays the same while pose estimation decreases smoothly - See Figure~\ref{F:KNN}-right for illustration). The results above consistently imply that the higher layers of CNN (expect FC8 which is task specific), even without any fine-tuning on the dataset, and even without any pose label optimization achieve representations that separate and highly preserve the view manifold structure. 

\begin{wrapfigure}{R}{0.7\textwidth}
\vspace{-4mm}
\centering
\includegraphics[width=\hsize,height=.35\hsize]{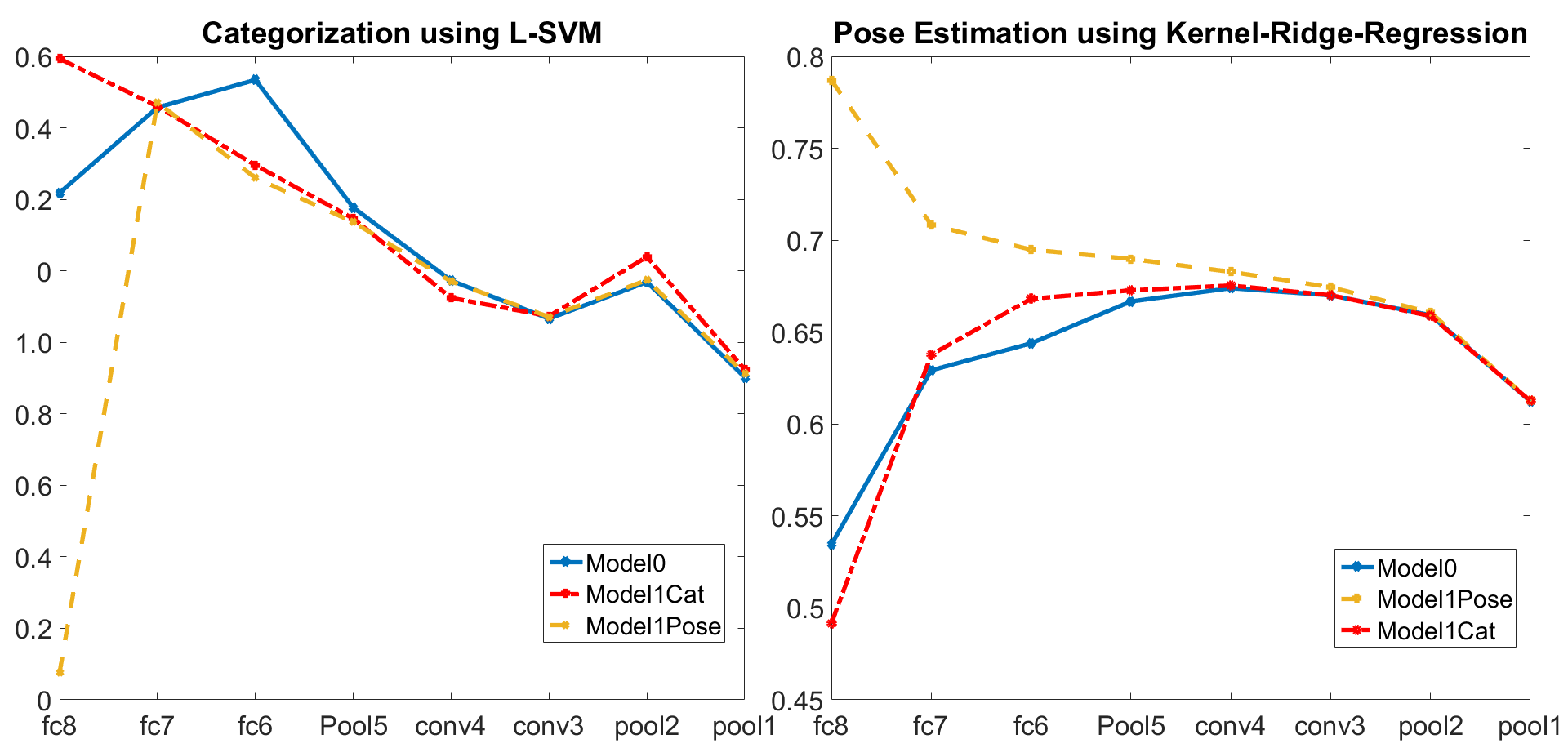}
\vspace{-4mm}
\caption{\footnotesize RGB-D: test performance of linear SVM category classification over the layers of different models (Left), and pose regression (Right).}
\vspace{-3mm}
\label{fig:SVMRidge1}
\end{wrapfigure}

The aforementioned conclusion is also confirmed by the test performance of Linear SVM and Kernel Regression in Figure~\ref{fig:SVMRidge1}, using RGBD dataset. 
In this experiment, the models are learned in train-split and the plots generated using test-split. 
Figure~\ref{fig:SVMRidge1} clearly shows the conflict in the representation of the pre-trained network (categorization increases and pose estimation decreases). Linear separability of category is almost monotonically increasing up to FC6.  Linear separability in  FC7 and FC8 is worse, which is expected as they are task specific (no fine-tuning). Surprisingly Pool1 features perform very bad, despite being the most general features (typically they show Gabor like features and color blobs). In contrast, for pose estimation, the performance increases as we go lower in the network up to Conv4 and then slightly decreases. This confirms our hypothesis that lower layers offer better feature encoding for pose estimation. It seems that Pool5 provides feature encoding that offer the best compromise in performance, which indicates that it is the best in compromising between the linear separation of categories and the preservation of the view-manifold structure. 

Surprisingly, the pose estimation results do not drop dramatically in FC6 and FC7. We can still estimate the pose (with accuracy around 63\%) from the representation at these layers, even without any training on pose labels. This highly suggests that the network preserves the view manifold structure to some degree. For examples taking the accuracy as probability at layer FC6, we can vaguely conclude that 90\% of the manifolds are linearly separable and 65\% are pose preserved (we are somewhere between hypotheses 1 and 2 at this layer). 

Table \ref{tab2_pascal} shows the quantitative results of our models on Pascal3D+ dataset. It also shows comparison against two previous methods \citep{zhang2015factorization} and \citep{xiang_pascal3d_2014}, using the two metrics $<45^\circ$ and $<22.5^\circ$ \footnote{Pose accuracy metrics are defined in \citep{zhang2015factorization,xiang_pascal3d_2014} and stated in Appendix~\ref{sec:metrics_pose}}. It is important to note that the comparison with \citep{xiang_pascal3d_2014} is unfair because they solve for detection and pose simultaneously while we solve for categorization and pose estimation. Model1 here outperforms both baselines (despite the unfair comparison with the latter approach). Quantitative results on RGBD dataset is presented in Appendix~\ref{sec:quant_rgbd}.



\begin{table}[th!]
\centering
\scriptsize
\begin{tabular}{|l|c|c|c|} 
\hline
\textbf{Approach} & \textbf{Categorization  $\%$} &\textbf{Pose (AAAI metric $\%$)}& \textbf{Pose (other metrics $\%$)}\\ 
\hline 
{\textit{Model0 (SVM/Kernel Regression)}} &FC6/FC7/FC8 & FC6/FC7/FC8&\\
 & 73.64/76.38/71.13 & 49.72/48.24/45.41 &\\ 
\hline 
{\textit{Model1 (SVM/Kernel Regression)}} & 74.65/79.25/84.12 & 54.41/54.07/60.31 &\\ 
\hline 
{\textit{Model0 NN}} & 60.05/69.89/61.26 & 61.11/61.38/60.32 &\\ 
\hline 
{\textit{Model1 NN}} & 73.50/77.30/83.07 & 65.87/66.07/70.54 &\\ 
\hline 
{\textit{Model1} (final prediction)} &  84.00 &   71.60 & 47.34($<$22.5), 61.30 ($<$45)  \\ \hline
~\citep{zhang2015factorization} &  - & - & 44.20 ($<$ 22.5), 59.00 ($<$45) \\  \hline
~\citep{xiang_pascal3d_2014} & -  & - & 15.6  ($<$22.5), 18.7 ($<$45)\\ 
\hline
\end{tabular} 
\caption{\footnotesize Pascal3D Performance computed for Model0 and Model1 using different classification techniques. Comparsion indicates that Model1 outperforms the baselines. 
}
\label{tab2_pascal}
\end{table}

\vspace{-5pt}
\section{Effect of Transfer Learning}
\label{S:FTAnalysis}
\vspace{-5pt}

In order to study the effect of fine-tuning  the network (transfer learning to a new dataset) on the representation we trained the following model (denoted as Model1). This architecture consists of two parallel CNNs: one with category output nodes (Model1-Cat), and one with binned pose output nodes (Model1-Pose). We used 34 and 11 category nodes for RGBD and Pascal3D datasets respectively; while we used 16 pose nodes for both datasets). The parameters of both CNNs were initialized by Model0 parameters up to FC7. The parameters connecting FC7 to the output nodes are randomly initialized on both networks and they are fine-tuned by minimizing the categorization loss for Model1-Cat and the pose loss for Model1-pose. The purpose of these architectures is to study the effect of fine-tuning when the category and pose are independently optimized. 

We applied all the measures described in Sec \ref{S:ManifoldAnalysis} to understand how the view manifolds will be affected after such tuning. The questions are: To what degree optimizing on category should damage the ability of the network to encode view manifolds. On the other hand, how  optimizing on pose should enhance that ability. 
Model1-Cat indicates the effect of optimizing on category, while Model1-Pose indicates the effect of optimizing on pose.

Fig~\ref{F:AnalysisPlots2} shows the five view manifold measures for the different layers of Model1(Cat/Pose), in comparison with Model0. In terms of data spread, from Fig~\ref{SV_sum1} shows that the spread at FC8 has doubled after fine tuning on pose  (Model1-Pose). Fig~\ref{EffectiveSV901} shows the fine tuning on category  (Model1-Cat) caused the view manifold subspace dimensionality to significantly reduce to 1, where it became totally view invariant. Optimizing on pose slightly enlarged the subspace dimensionality ({\emph{i.e,}} become better) at FC8 and FC7. Fig~\ref{kta1} clearly shows the significant improvement achieved by fine tuning on pose, where the alignment of FC8 jumped to close to 0.9  from about 0.78, while fine tuning on category reduces the alignment of FC8 to close to 0.65. Similar behavior is also apparent in the KPLS ratio for FC8 and FC7 (sup-mat). 

One very surprising result is that optimizing on pose makes the pose KTA alignment worse at the lower layers, while optimizing on category makes the pose alignment better compared to model0. In fact, although optimizing on pose significantly helps aligning FC8 with pose labels, Pool5 still achieves the best KTA alignment and the least regression reconstruction error. 
The regression reconstruction error in Fig~\ref{regression_err1} clearly shows significant improvement in the representation of FC8 and FC7 to preserve the view manifold. One surprising finding from these plots is that the representation of FC6 becomes worse after fine tuning for both pose and category. Fig~\ref{tps_poly1} indicates that the deformation of the view manifold is reduced as a result of fine tuning on pose (larger rcond number), while it increases as a result of fine tuning on category.

On the global object-view manifold structure, we notice from Figures \ref{fig:SVMRidge1} some intuitive behavior at FC8. Basically optimizing on pose reduces the linear separability and increases the view manifold preservation (moves the representation towards hypothesis 0). In contrast, optimizing the category significantly improves the linear separability at FC8, however, interestingly, it only slightly reduces the pose estimation performance to be slightly less than 50\%. Combining this conclusion with the observation from Fig~\ref{EffectiveSV901}, that the view manifold subspace dimensionality reduces to 1, this implies that optimizing on category collapses the view manifolds to a line, but they are not totally degenerate. 
What is less obvious is the effect of fine tuning on the lower layers than FC8. Surprisingly, optimizing on pose did not affect the linear separability of FC7. Another very interesting observation is that optimizing on category actually improves the pose estimation slightly at the FC7, FC6, and Pool5; and did not reduce it at lower layers. This implies that fine tuning by optimizing on category only improved the internal view manifold preservation at the network, even without any pose labels. 

\section{Conclusions}


In this paper we present an in-depth analysis and discussion of the view-invariant properties of CNNs. We proposed a methodology to analyze individual instance's view manifolds, as well as the global object-view manifold. We applied the methodology on a pre-trained CNN, as well as two fine-tuned CNNs, one optimized for category and one for pose. We performed the analysis based on two multi view datasets (RGBD and Pascal3D+). 
Applications on both datasets give consistent conclusions.

Based on the proposed methodology and the datasets, we analyzed the layers of the pre-trained and fine-tuned CNNs.
There are several findings from our analysis that are detailed throughout the paper, some of them are intuitive and some are surprising. We find that a pre-trained network captures representations that highly preserve the manifold structure at most of the network layers, including the fully connected layers, except the final layer. Although the model is pre-trained on ImageNet, not a densely sampled multi-view dataset, still, the layers have the capacity to encode view manifold structure.
It is clear from the analysis that, except of the last layer, the representation tries to achieve view invariance by separating individual instances' view manifolds while preserving them, instead of collapsing the view manifolds to degenerate representations. This is violated at the last layer which enforces view invariance.

{\emph Overall, our analysis using linear SVM, kernel regression, KNN, combined with the manifold analysis, makes us believe that CNN is a model that simulate the manifold flattening hypothesis of ~\cite{dicarlo2007untangling} even without training on multi-view dataset and without involving pose labels in the objective's loss.}

Another interesting finding is that Pool 5 offers a feature space where the manifold structure is still preserved to the best degree. Pool 5 shows better representation for the view-manifold than early layers like Pool1. We hypothesize that this is because Pool5 has better translation and rotation invariant properties, which enhance the representation of the view manifold encoding. 

We also showed the effect of fine-tuning the network on multi-view datasets, which can achieve very good pose estimation performance. In this paper we only studied the effect of independent pose and category loss optimization. Optimizing on category achieves view invariance at the very last fully connected layers; interestingly it enhances the viewpoint preservation at earlier layers.  We also find that fine-tuning mainly affects the higher layers and rarely affects the lower layers. 

In this work our goal is not to propose any new architecture or algorithm to compete with the state of the art in pose estimation. However, the proposed methodology can be used to guide deep network design for solving several tasks. To show that and based on the analysis and the conclusions of this paper, we introduced and studied in \citep{dp2main_Elhoseiny2015} several variants of CNN architectures for joint learning of pose and category, which outperform the state of the art . 
We keep these results as a guide to the reviewers, without distracting the reader from our main goal. 


\textbf{Acknowledgment: } This work is funded by NSF-USA award \# 1409683.

%
%
%
%
%
 


{
\bibliography{DP1-main}
\bibliographystyle{iclr2016_conference}
}

\clearpage

\section*{Appendix}
\appendix

\section{Pose and Categorization Performance on RGBD and Pascal3D datasets}
\label{sec:metrics_pose}
The two metrics $<22.5$ and $<45$ are the percentages of test samples that satisfy $AE<22.5^{\circ}$ and $AE<45^{\circ}$, respectively where the Absolute Error (AE) is $AE= | Estimated Angle - Ground Truth|$). 
The AAAI pose metric is defined as

\begin{equation}
\Delta (\theta_i,\theta_j) = min{(|\theta_i - \theta_j|,2\pi - |\theta_i - \theta_j|)} / \pi
\end{equation}


\section{Quantitative results for RGBD Dataset}
\label{sec:quant_rgbd}
Table~\ref{tab1_rgbd} shows qualitative ressult of Kernel-SVM Regression on different layers features in Model0 and Model1. Comparing their results with images HOG features. Comparison to state of Art using this model is not possible since we work on the wellposed objects split which is not explored by any other work.

\begin{table}
	\scriptsize
	\centering
	\begin{tabular}{|l|c|c|}
		\hline 
		\textbf{Approach} & \textbf{Categorization} $\%$ & \textbf{Pose} $\%$ \\ 
		\hline 
		HOG (SVM/Kernel Regression) & 80.26 & 27.95 (AAAI) \\ 
		\hline 
		Model0 (SVM/Kernel Regression) on conv4 & 58.64 & 67.39 (AAAI)\\ 
		\hline 
		Model0 (SVM/Kernel Regression) on FC6 & 86.71 & 64.39 (AAAI) \\ 
		\hline 
		Model1 & 89.63 & 81.21 (AAAI), 69.58 ($<$ 22.5), 81.09 ($<$ 45) \\ 
		\hline 
	\end{tabular} 
	\caption{\footnotesize RGBD Dataset Results for HOG, Model0 and Model1.}
	\label{tab1_rgbd}
\end{table}

\section{Synthetic Data Analysis}
\label{sec:synthetic_dataset}
In this work, we have explored many different measurements and filter them out to use only those that expose the correct properties of the view-manifolds. Besides the intuitive reasoning that we provided for choosing the measurements, in this section, we show empirical results to quantify efficiency of the chosen measurements. 

To this end, we synthesized a set of well designed view-manifolds. Analyzing these manifolds is intended to identify the robust and informative set of measurements to be used in further analysis. 
To be qualified for comparing different manifolds, the synthesized manifolds is designed to encode interesting properties of any view-manifold such as:
\begin{itemize}
\item Dimensionality (of the Euclidean space where the manifold lives)
\item Sparsity of the manifold
\item Smoothness of the manifold
\item Deformation of the manifold w.r.t the \textit{view-circle}
\item Variance of data-points
\end{itemize}

Recall, The view-circle is a view-manifold, where all the viewpoints form a perfect circle and the object is assumed to be located at the center of this circle.
In the rest of this section, we list detailed description of the synthetic manifolds. Then, we use them to analyze the selected measurements.

\begin{figure*}[ht!]
	\centering
	\begin{subfigure}[b]{0.3\columnwidth}
		\tiny
		\caption{Sinosoidal Surface}
		\centering
		\includegraphics[width=\textwidth,height=0.75\textwidth]{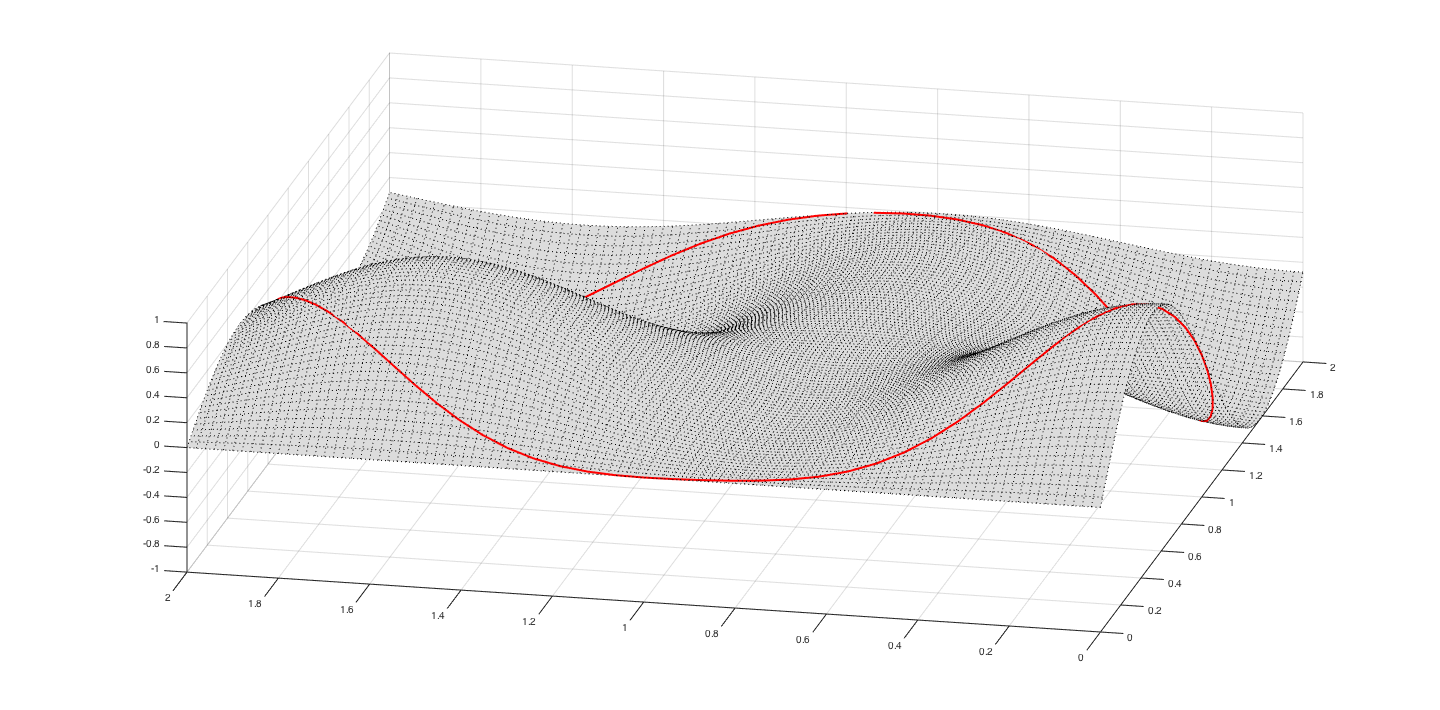}
		\label{SynthManifold3}
	\end{subfigure} 
	\begin{subfigure}[b]{0.3\columnwidth}
		\tiny
		\caption{Circle projected on $S_2$}
		\centering
		\includegraphics[width=\textwidth]{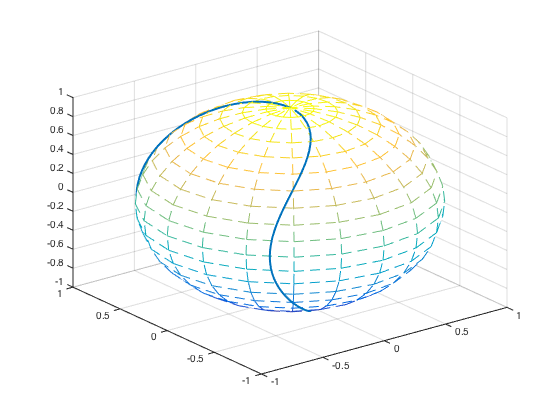}
		\label{SynthManifold4}
	\end{subfigure} 
	\begin{subfigure}[b]{0.3\columnwidth}
		\tiny
		\caption{Circle projected on $S_2$}
		\centering
		\includegraphics[width=\textwidth]{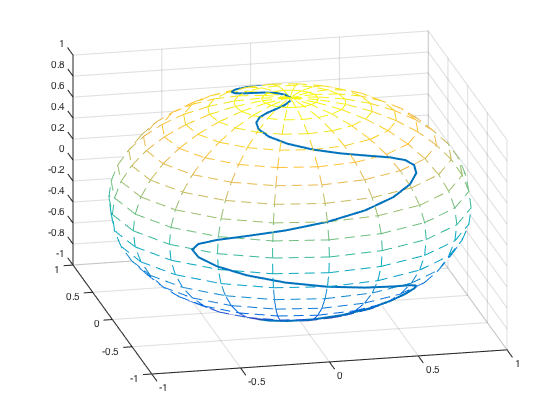}
		\label{SynthManifold6}
	\end{subfigure} 
	\caption{Manifold Visualization}
	\label{F:manifolds_types}
\end{figure*}

\subsection{Dataset Description}
As in Figure~\ref{F:manifolds_types}, manifolds in this dataset can be categories as: 

\begin{itemize}
\item Circle Orthogonally projected to  high-dimensional subspaces (Manifold sets 1 and 2)
\item Unit circle projected to a nonlinear surface (manifold 3)
\item Unit circle projected to $3D$-Sphere with radius $r$ ($S^r_2$) (sets 4 an 5)
\item Nonlinear smooth curve projected on $S^r_2$ (set 6)
\item Discontinuous smooth curve projected on $S^r_2$ (set 7)
\item Random manifolds (sets 8 and 9 ) 
\item Collapsed manifolds in a single point or very small region (set 10). 
\end{itemize}

The manifolds are described using the dimensionality ($d$), sparsity ($s=\frac{n}{d}$) and $n$, the number of points representing the view-manifold, smoothness, deformation w.r.t the view-circle.

Let the view-manifold be parameterized by the single dimensional variable. Let $S$ is the two dimensional representation of the unit circle. $S=\{(cos(t),sin(t))| t=\{0,\frac{2\pi}{n},\frac{4\pi}{n}, ..., \frac{2(n-1)\pi}{n} \}\}$. 
For each view-manifold ($\cal{M}$), we generated $n$ points in a $d$-Dim space.
\begin{itemize}
\item \textbf{Perfect view-circle in high-dimensional space }
%

\begin{itemize}
\item Manifold 1: $n=100$, $d\in\{10,300,600,900,1200,1500,1800\}$, therefore, the sparsity varies from very dense ($s=10$) to very sparse ($s=1/20$) 
\item Manifold 2: $d=500$,  $n\in\{50,150,250,350,450,550,650,750\}$, therefore, the manifold varies from very sparse ($s=1/10$) to dense ($s=1.5$) 
\end{itemize}

\item \textbf{View-circle projected nonlinearly to Sinusoidal Surface}. The manifold has $n=100$ points and live in $d=3$-Dim space, so it is  very dense $s=33.33$ 
To project the view-circle on this surface we follow these steps:
\begin{itemize}
\item Let $fn$ be the projection function on the surface, 
\[fn(x,y) = sin(3x)cos(2y)^2\]
\item The projected manifold $Z$ is defined by
\[Z=\{(x,y,fn(x,y))|(x,y)\in S\}\]
\end{itemize}

\begin{itemize}
\item Manifold 3 represents this type of manifolds in our dataset.
\end{itemize}

\item \textbf{Dense view-circle projected nonlinearly to $S^r_2$, with $r\in\{1,50,100,150\}$, $d=3$, $n=100$\   ($s=33.33$)}

To project the view-circle on $S_2^r$, we use the following projection function
\[f(\theta,\phi)=(sin(\phi)cos(\theta),sin(\phi)sin(\theta),cos(\phi))\]
Where
\[\theta \in \{0,\frac{2\pi}{n},\frac{4\pi}{n}, ..., \frac{2(n-1)\pi}{n} \}\] 

\begin{itemize}
\item Manifold 4: Slightly deformed manifold, Figure~\ref{SynthManifold4}
\[\phi = \frac{\pi}{4}sin(\theta)+\frac{\pi}{2};\forall \theta\]
\item Manifold 5: Slightly deformed manifold with added Gaussian noise with $\mu=0$ mean $\sigma=0.01$. $\theta$ and $\phi$ as in Manifold 4.
\item Manifold 6: Highly deformed manifold, Figure~\ref{SynthManifold6}
\[\phi = \frac{\pi}{4}sin(5\theta)+\frac{\pi}{2};\forall \theta\]
\item Manifold 7: Highly deformed and broken/discontinuous manifold. 
\[\phi = \frac{\pi}{4}tan(0.75\theta)+\frac{\pi}{2};\forall \theta\]
\end{itemize}

\item \textbf{Random manifold with independent dimensions}
\begin{itemize}
\item Manifold 8: Uniform random points with $d \in \{10,100,500,1000,4000\}, n=100$, therefore, the sparsity varies from very dense ($s=10$) to very sparse ($s=1/100$)
\item Manifold 9: Normal random points has been generated with $d=100, n\in\{20,40,..., 200\}$, therefore, sparsity varies from very sparse ($s=1/10$) to dense ($s=2$)
\end{itemize}
\item \textbf{Collapsed Manifold with random noise}
\begin{itemize}
\item Manifold 10: Portion of the points ($m=n/4$), in this manifold, have been generated by Gaussian Random with $\mu=0$, $\sigma=0.01$, therefore, the rest are a copied version of this portion $d \in \{10,100,500,1000,4000\}, n=100$
\end{itemize}

\end{itemize}
%


\subsection{Analysis}
\begin{figure}[ht!]
\centering
\begin{subfigure}[b]{0.32\columnwidth}
  \caption{\scriptsize Nuclear Norm}
  \centering
    \includegraphics[width=\textwidth,height=0.75\textwidth]{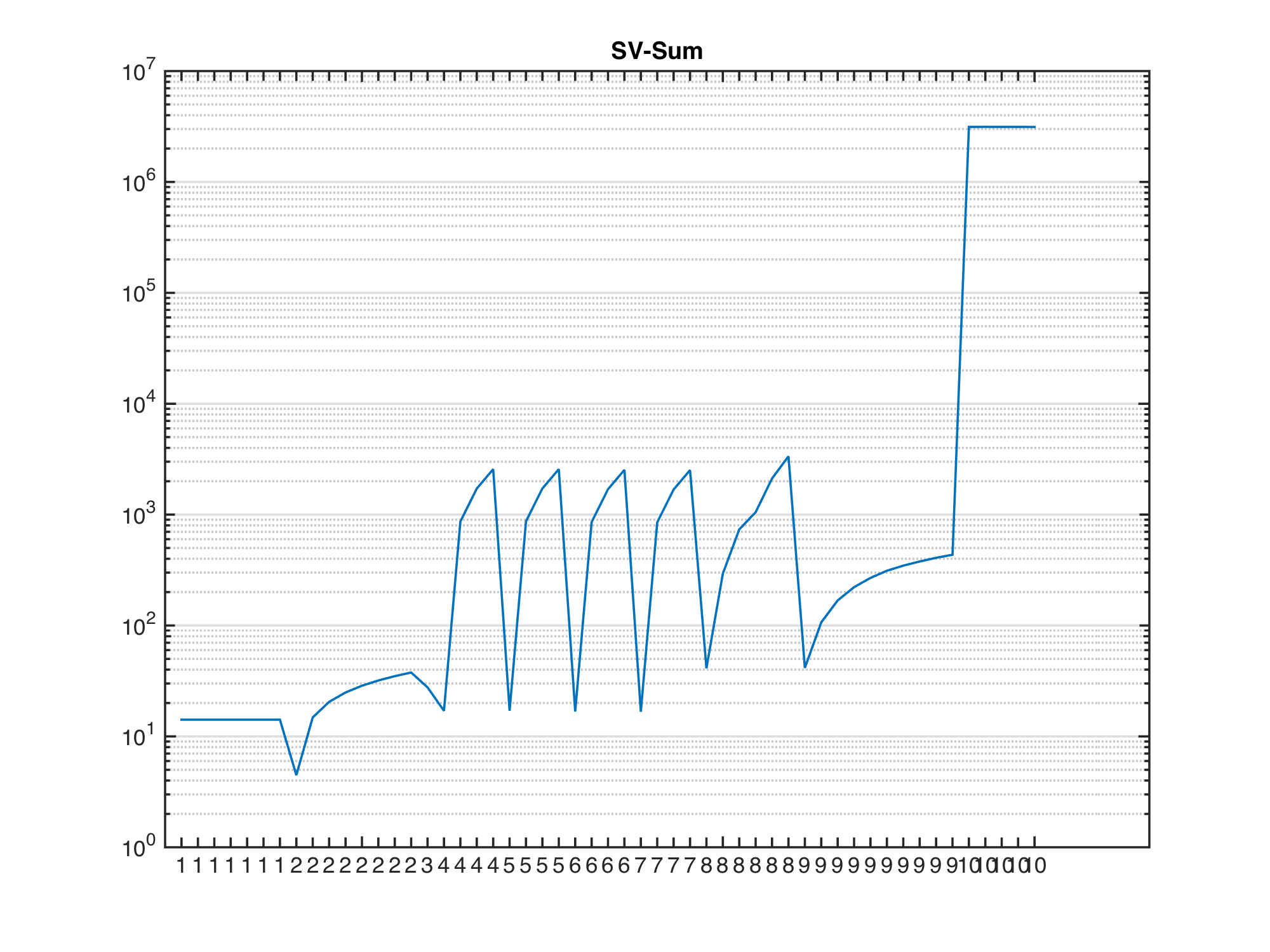}
        \label{synth:SV_sum}
\end{subfigure} 
\begin{subfigure}[b]{0.32\columnwidth}
  \caption{\scriptsize Effective $90\%$ SV's}
  \centering
    \includegraphics[width=\textwidth]{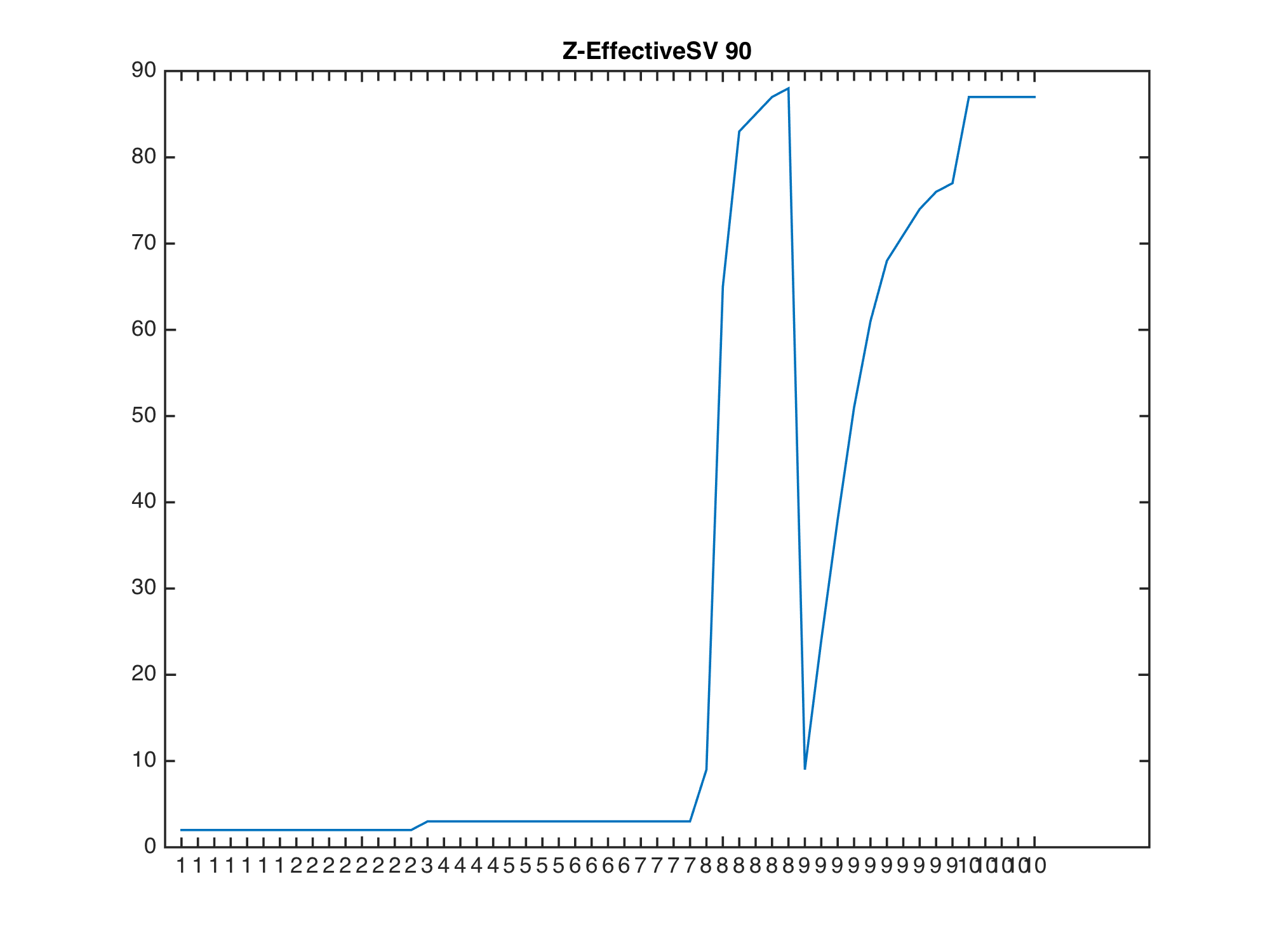}
        \label{synth:EffectiveSV90}
\end{subfigure} 
\begin{subfigure}[b]{0.32\columnwidth}
  \caption{\scriptsize KTA}
  \centering
    \includegraphics[width=\textwidth]{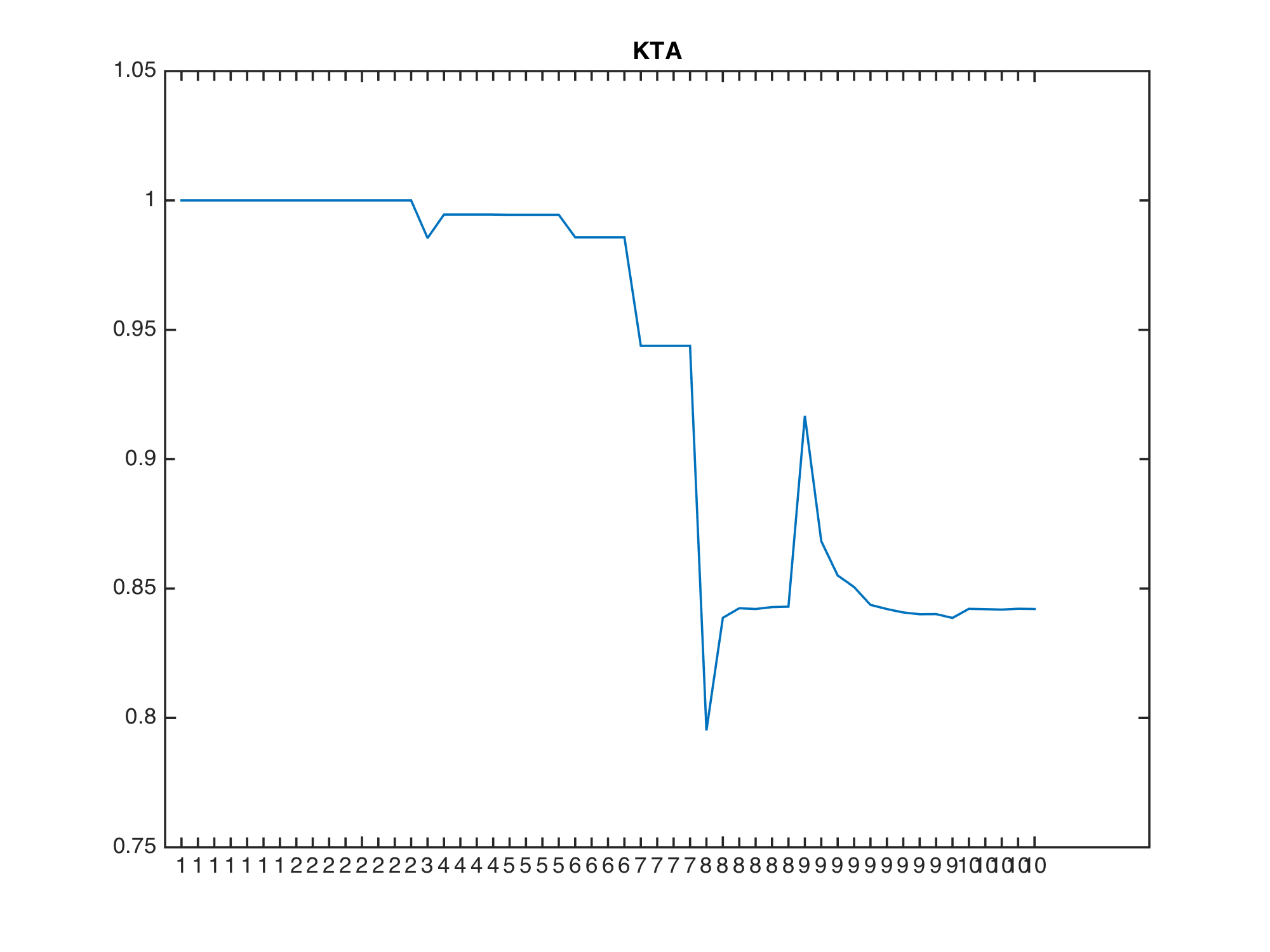}
    \label{synth:kta}
\end{subfigure}
\begin{subfigure}[b]{0.32\columnwidth}
  \caption{\scriptsize HSIC}
  \centering
    \includegraphics[width=\textwidth]{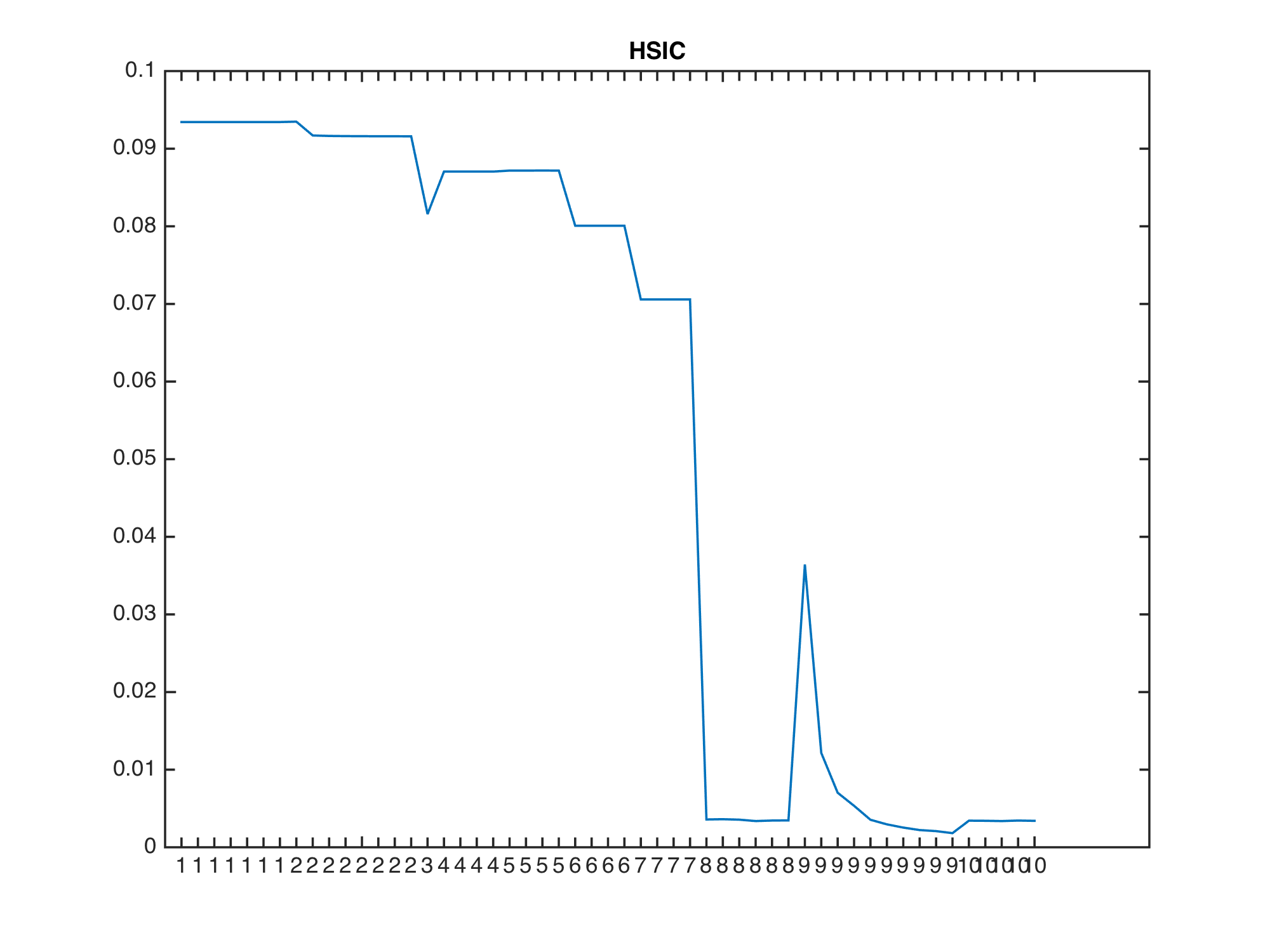}
    \label{synth:hsic}
\end{subfigure} 
\begin{subfigure}[b]{0.32\columnwidth}
  \caption{\scriptsize KPLS-Regression error}
  \centering
    \includegraphics[width=\textwidth]{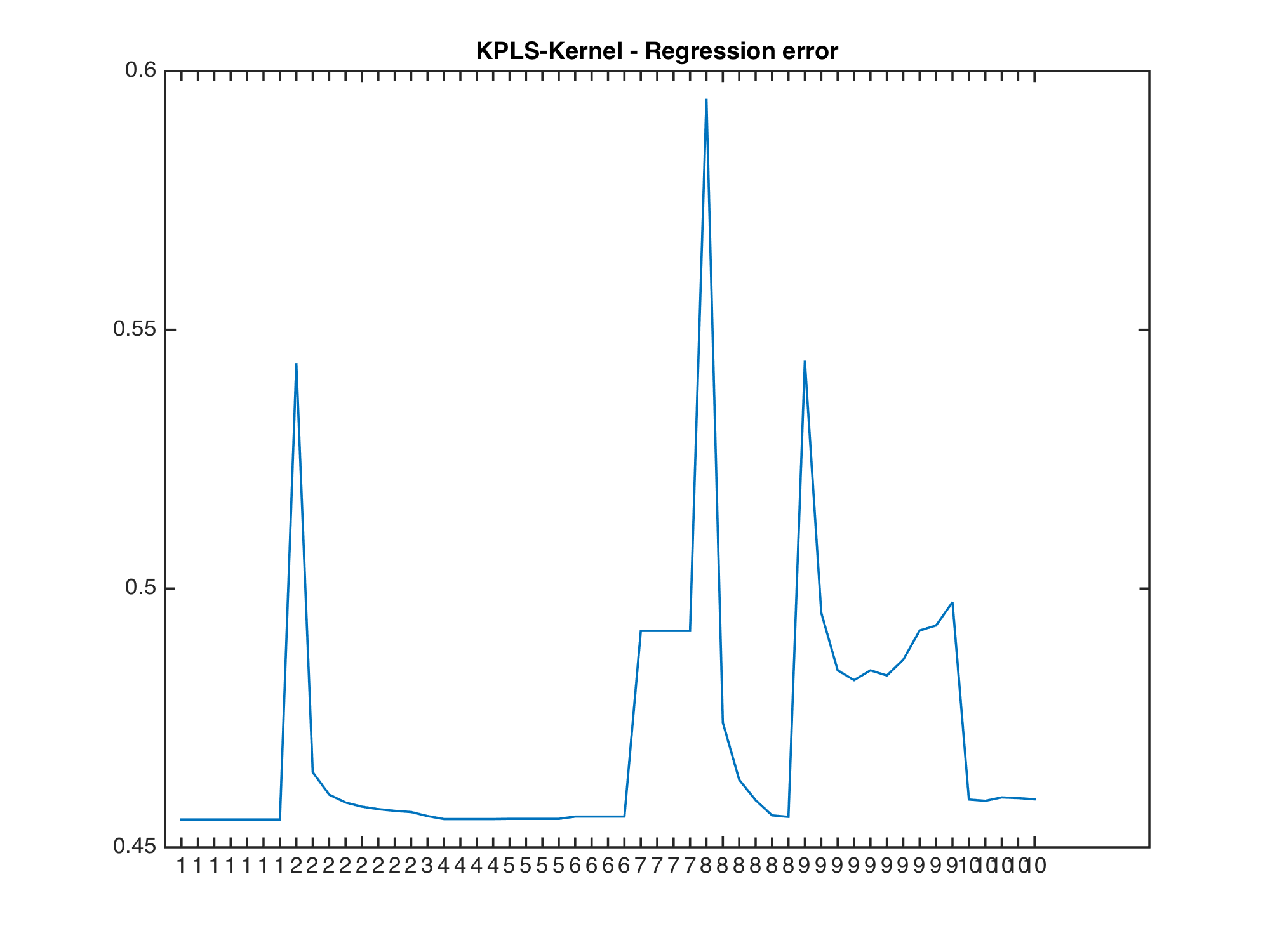}
    \label{synth:regression_err}
\end{subfigure}
\begin{subfigure}[b]{0.32\columnwidth}
  \caption{\scriptsize KPLS-$\frac{norm(G_d)}{norm(G_0)}$}
  \centering
    \includegraphics[width=\textwidth]{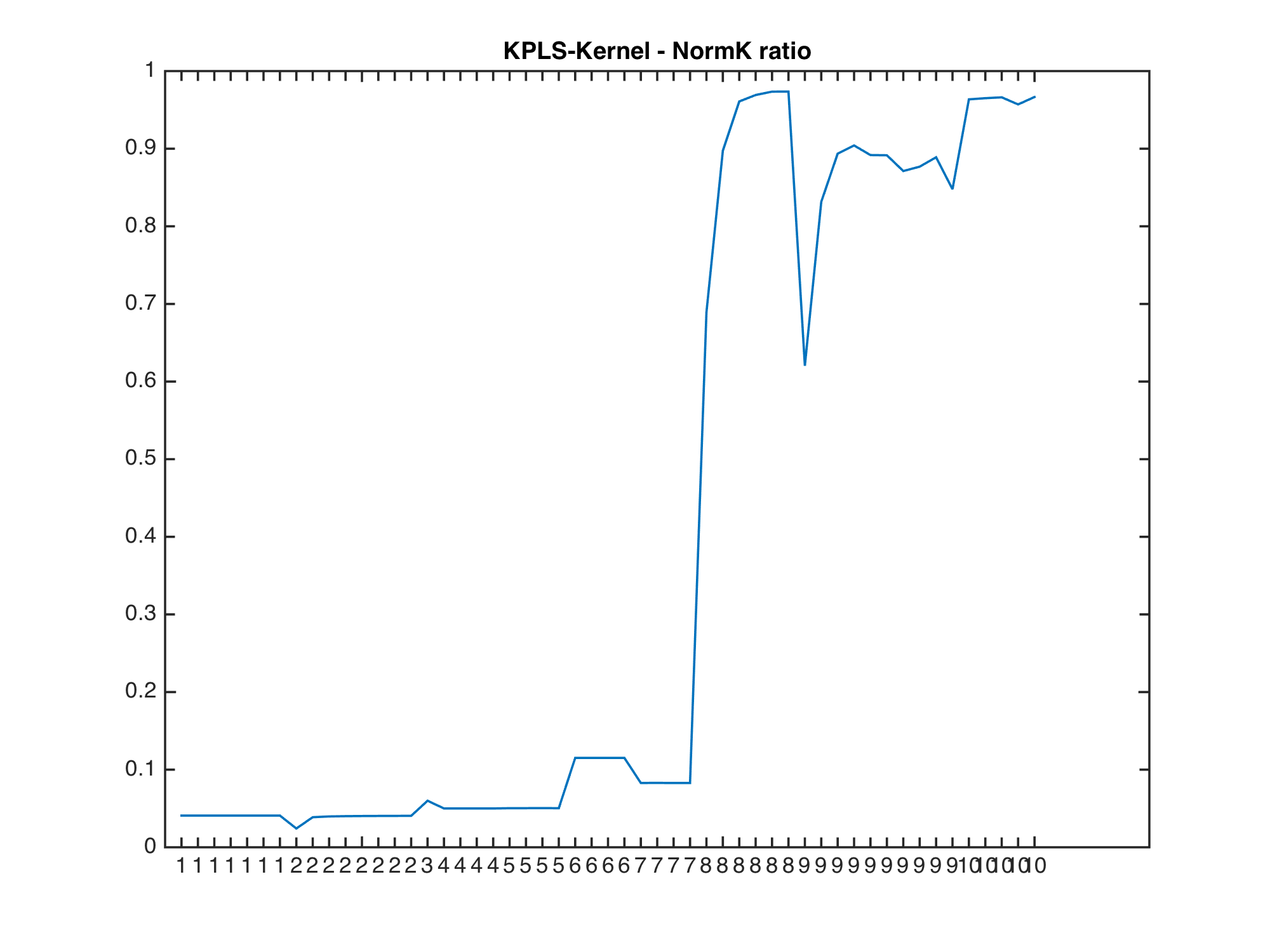}
    \label{synth:normk_ratio}
\end{subfigure}
\begin{subfigure}[b]{0.32\columnwidth}
  \caption{\scriptsize TPS-$RCond(CF-poly)$}
  \centering
    \includegraphics[width=\textwidth,height=0.75\textwidth]{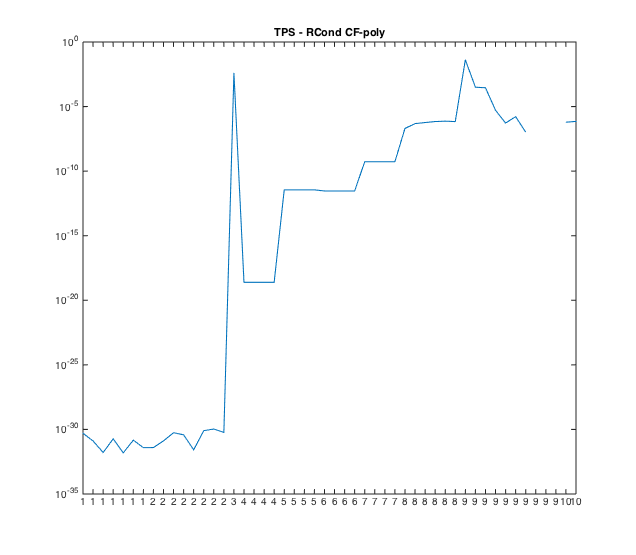}
    \label{synth:tps_poly}
\end{subfigure}
\begin{subfigure}[b]{0.32\columnwidth}
  \caption{\scriptsize TPS-$RCond(CF-nonPoly)$}
  \centering
    \includegraphics[width=\textwidth,height=0.75\textwidth]{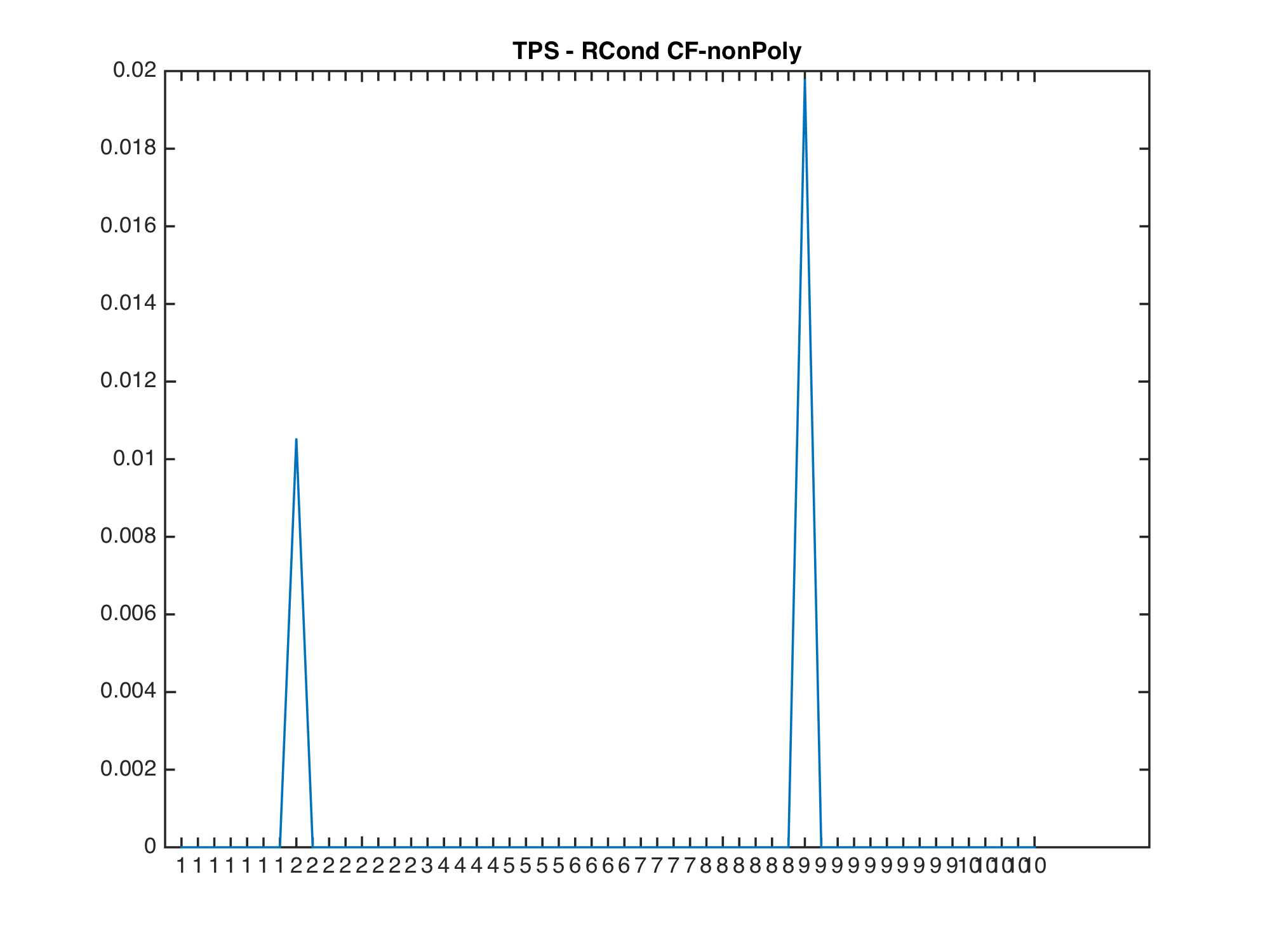}
    \label{synth:tps_nonpoly}
\end{subfigure}
\caption{Measurement analysis for the synthetic manifolds. Every figure shows single measurement. X-axis is labeled by the manifold category number.}
\label{SyntheticMeasurements}
\end{figure}

Recall, the objective of using the synthetic data is to verify the efficiency of selected measurements. Figure~\ref{SyntheticMeasurements} shows the results of applying the measurements to the synthetic-data. 
Figure~\ref{synth:SV_sum} shows the Nuclear Norm (defined in the main paper) for all manifolds. This figure shows the variability between the manifolds in the variance. For the set of manifolds 4-7, projecting the view-circle onto sphere with different sizes affects the variance of the points. Encoding different Nuclear Norm is subjected to discover the measurements that are sensitive to the data variance.

From Figure~\ref{synth:EffectiveSV90}, defined in the main paper, we can see the effective dimensions for each manifold. Manifolds 1 and 2 have two effective dimensions. Manifolds 3-7 has three effective dimensions. Since the points in Manifolds 8-10 are generated randomly so they have maximum rank. 


The kernel alignment measures: KTA (Figure~\ref{synth:kta}) and HISIC (Figure~\ref{synth:hsic}) measure the correlation between the view-manifold and the view-circle. These two figures show significant better alignment of the view-manifold of sets 1-6 
than the alignment of the random manifolds. Since Hilbert-Schmidt Independence Criterion (HSIC) \cite{Gretton_2005} does not add any information more than KTA. We select the KTA measurement because it exposes absolute alignment confidence for the manifolds 1 and 2.

KPLS-regression Error is shown in Figure~\ref{synth:regression_err}. Dispite the vast variability of variance and dimensionality, this measure is consistent and gives small value for all smooth manifold. This measure can also detect the collapsing manifolds, since it gives very large error value. 

As we mentioned in Section~\ref{sec:moreMeaurements}, that using both measurements KPLS-Regression Error and KPLS-Norm Ratio gives more robust conclusion about the manifold. Fig~\ref{synth:normk_ratio} shows a clear trend, since it gives significant high values for 
random manifolds. This is because, the subspace of the random points covers the entire space. When $\frac{norm(G_d)}{norm(G_0)}\equiv 1$, this means that the firt $d$ components extracted from $G_0$ are far from being principal. If they are pricipal components, they would change the energy of the matrix $G$ significantly.
On the other side, the Effecive dimensionality of the smooth manifolds 1-6 is $D\leq 3$, which make the limit $d>D$. That is why the ratio $\frac{norm(G_d)}{norm(G_0)}\ll$ because we have extracted all the pricipal components of those manifolds. That is why KPLS-Regression Error for these manifolds is very small.

As mentioned in the main paper, TPS-lineairty measure ($TPS-RCond(CF-Poly)$) scores on the stability of the polynomial mapping from the points on the view-circle and the points on the view-manifold. Fig~\ref{synth:tps_poly} shows perfect scoring for Manifolds 1 and 2. 
Combining this figure with Fig~\ref{synth:tps_nonpoly} gives a complete impression about the mapping stability (Polynomial and Non-Polynomial). However, the range of the values of TPS-nonlinearity measure ($TPS-RCond(CF-nonPoly)$) is in $ BigO(10^{-8})$, which decrease its robustness.

%
%
%
%
%
 



\section{More on KPLS and TPS related Measurements}
\label{sec:moreMeaurements}

We define and show the results of more measurements such as KPLS-Norm Ratio and TPS-nonPolynomial. We use here the same notations and definitions stated in Section 4 in the main paper.

\medskip
\noindent{\emph 1) KPLS-Norm Ratio:}  
Kernel Partial Least Squares (KPLS) \cite{Rosipal_kpls_2002} is a supervised regression method. KPLS iteratively extracts a set of principal components of the input kernel that are most correlated with the output. 
While KPCA extracts the principal components (PCs) of the kernel of the input data to maximize the variance of the output space, KPLS extracts the PCs of the kernel of the input data that maximize the correlation with the output data.
We use KPLS to map the affinity matrix of the transformed view-manifold  ({\emph view-kernel}) to the circle affinity matrix ({\emph circle-kernel}). Following the convention of the main paper, let the view-kernel is denoted by  $\mathbf{K}^l$, and the circle-kernel is denoted by $\mathbf{K}^\circ$ ({\emph The subscript $n$ is removed to simplify the notation}).
We limit the number of extracted PCs to $d$ , where $d\ll N$ and $N$ is the dimensionality of the input kernel (in this work, we use $d=5$).
More specifically, KPLS maps the rows of $\mathbf{K}^l$ to the rows of $\mathbf{K}^\circ$. So that
\begin{equation}
\label{eq:kpls_regression}
\hat{\mathbf{K}}^\circ= \mathbf{G_0U}(\mathbf{T^\top G_0U})^{-1}\mathbf{T}^\top \mathbf{K}^l
\end{equation}
Where the set of extracted PCs are the columns of the matrix $T_{N\times d}$, $U_{N\times d}$ is auxiliary matrix, and the Gram-matrix $\mathbf{G_0}$ is defined by
\begin{equation}
\mathbf{G}_0= \frac{{\mathbf{K}^l} {\mathbf{K}^l}^\top}{b b^\top}
\label{eq:G_0}
\end{equation}
Where $b\in \mathbb{R}^N$ 
, so that $b(i)$ is the Frobenius norm of the $i$-th row of $\mathbf{K}^l$. 
Based on the mapping in Eq~\ref{eq:kpls_regression}, we extract two measurements: 

First: {\emph KPLS-Regression Error ($\delta$)} which measures geometric deformation of the generated output image of view-kernel in the circle-kernel space ($\hat{\mathbf{K}}^\circ$ with respect to the the circle-kernel (${\mathbf{K}}^\circ$) and the ). One choice for measuring this is
\[\delta(\hat{\mathbf{K}}^\circ , {\mathbf{K}}^\circ) = 1- KTA(\hat{\mathbf{K}}^\circ , {\mathbf{K}}^\circ) \]
Where KTA stands for Kernel Target Alignment (stated in Equation $2$ in the main paper).
The Regression error measures the reconstruction error of the circle-kernel from the view-kernel.
%

Second, KPLS-NormK Ratio ($\frac{\|G_d\|_{F}}{\|G_0\|_{F}}$) measure the residual energy after extracting the first $d$-PC's. Where $G_d$ is the residual of $G_0$ after $d$-iterations. 
The intuition behind this measure is that the larger the ratio $\frac{\|G_d\|_{F}}{\|G_0\|_{F}}$, this means that the view-manifold has more than $d$-PC's correlated with the circle-kernel. 

\begin{figure}[h!]
\centering
\begin{subfigure}[b]{0.49\columnwidth}
  \caption{\scriptsize KPLS-Regression Error ($\delta$)}
  \centering
    \includegraphics[width=\textwidth, height=.85\textwidth]{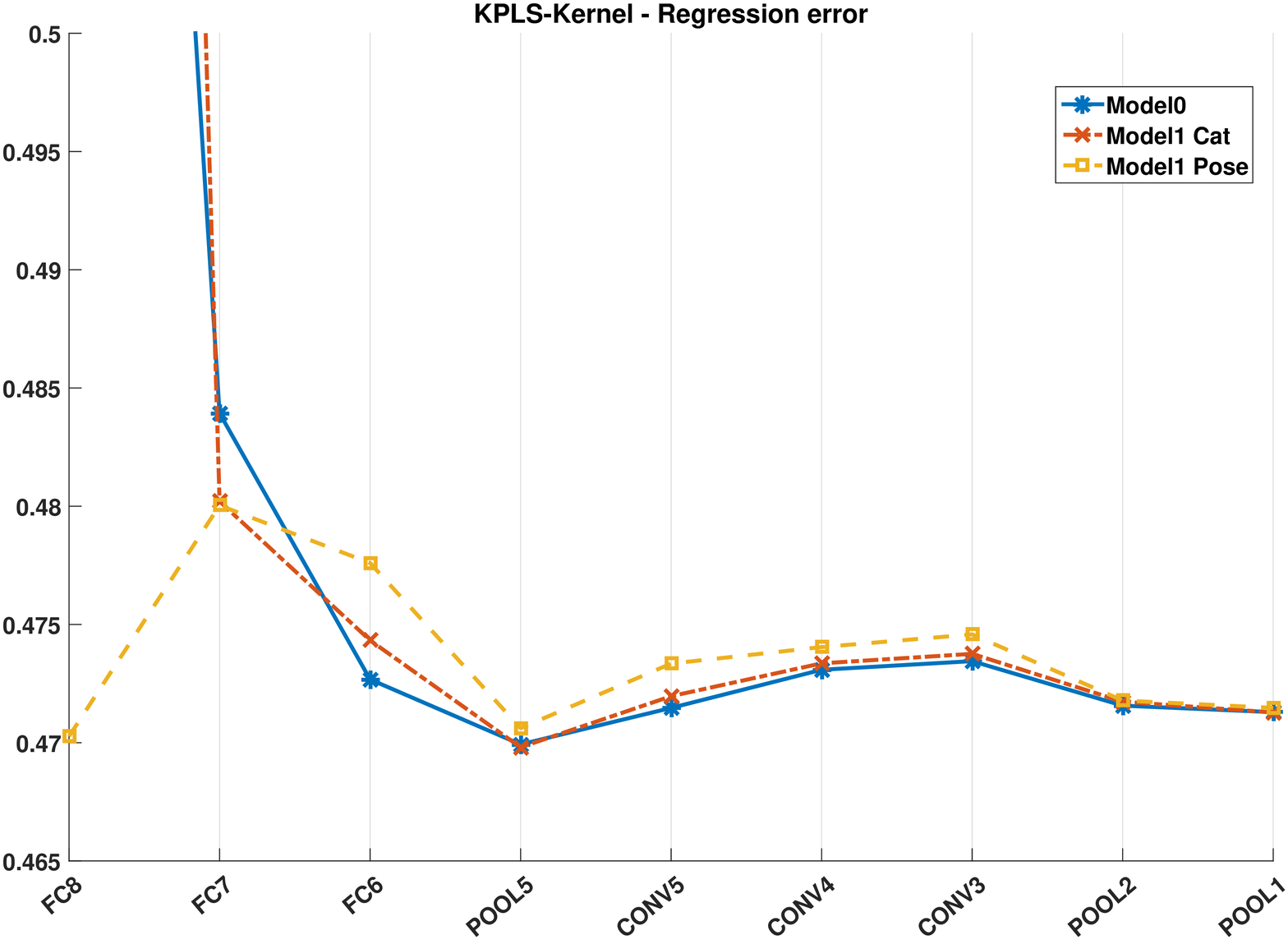}
    \label{regression_err}
\end{subfigure}
\begin{subfigure}[b]{0.49\columnwidth}
  \caption{\scriptsize KPLS-Norm Ratio ($\frac{norm(G_d)}{norm(G_0)}$)}
  \centering
    \includegraphics[width=\textwidth, height=.85\textwidth]{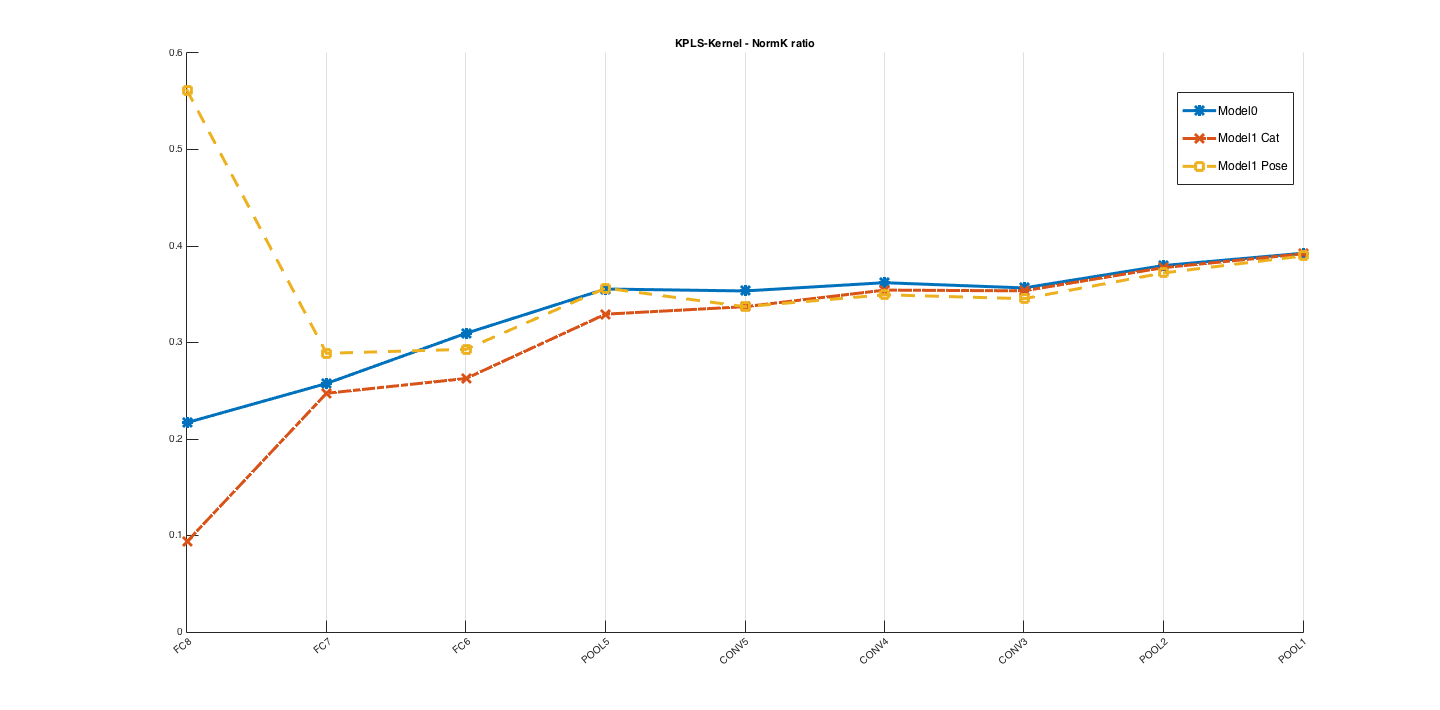}
    \label{normk_ratio}
\end{subfigure}
\begin{subfigure}[b]{0.49\columnwidth}
  \caption{\scriptsize TPS-$RCond(CF-poly)$}
  \centering
    \includegraphics[width=\textwidth, height=.85\textwidth]{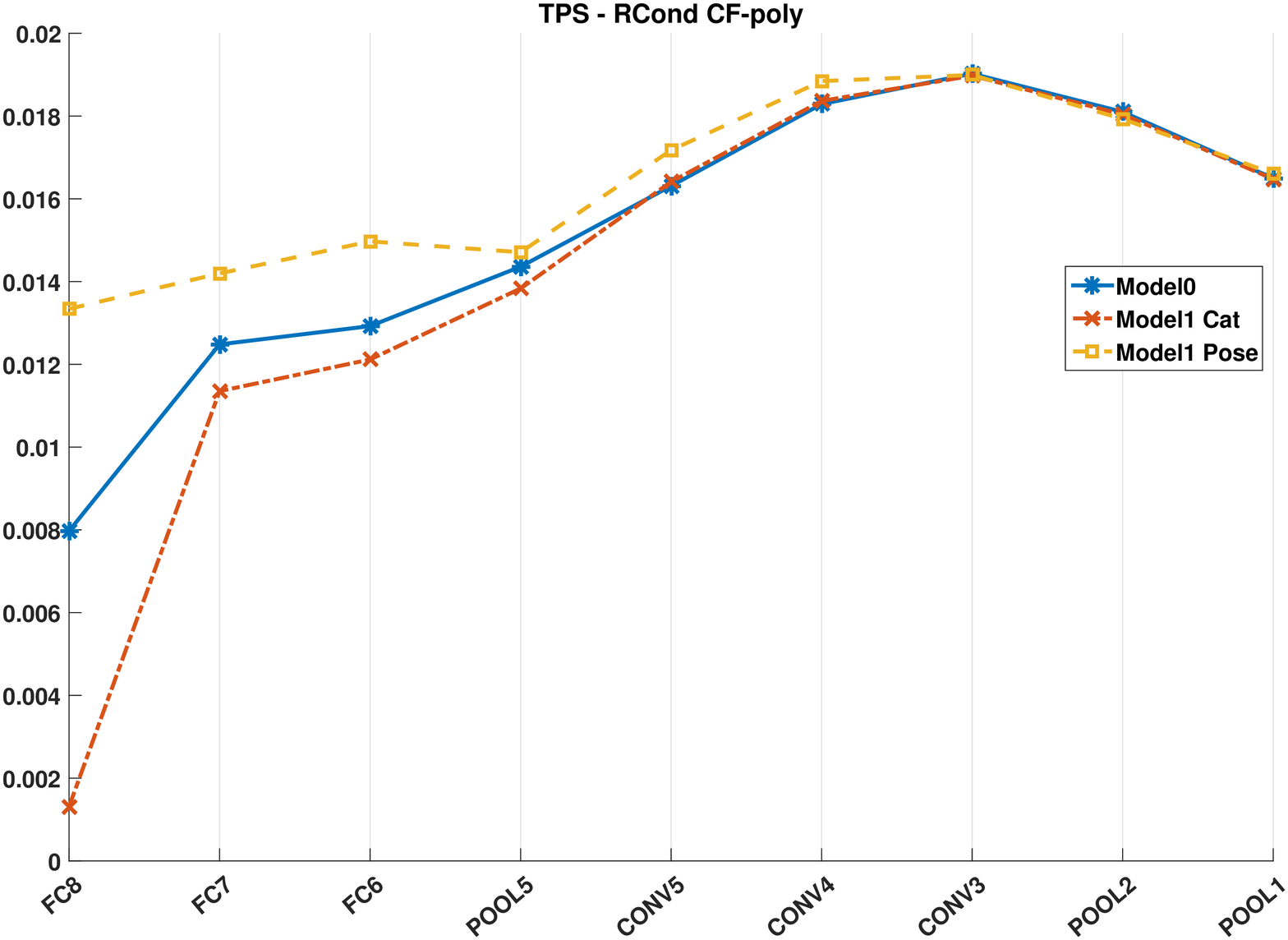}
    \label{tps_poly}
\end{subfigure}
\begin{subfigure}[b]{0.49\columnwidth}
  \caption{\scriptsize TPS-$RCond(CF-nonPoly)$}
  \centering
    \includegraphics[width=\textwidth,height=0.85\textwidth]{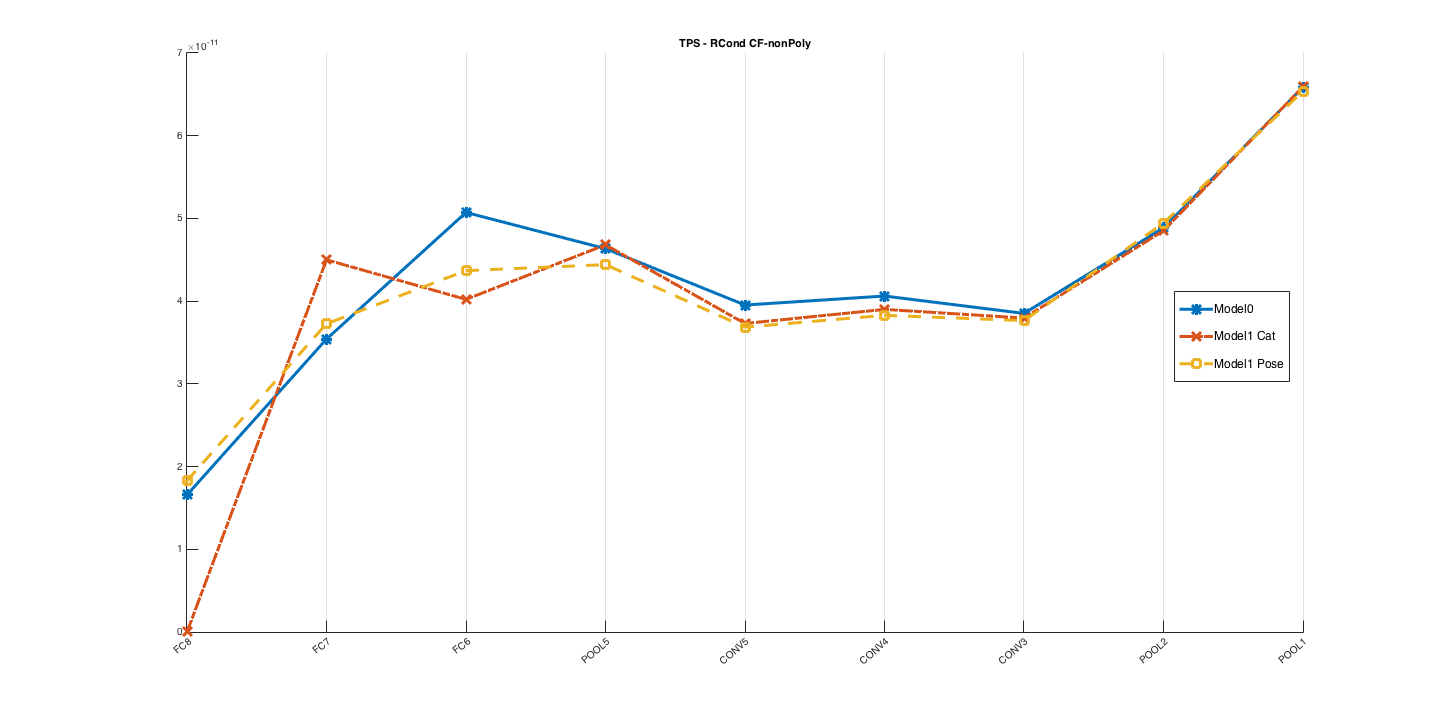}
    \label{tps_cf}
\end{subfigure}
\caption{Measurement analysis for the view-manifold in RGBD dataset based on features extracted from different layers of several CNN models. Every figure shows single measurement. Multiple lines is for different CNN model. X-axis is labeled by the layers.}
\end{figure}

While KPLS-regression Error is self-explanatory (this measure presented in the main paper), using the two KPLS measurements together gives more precise view on the correlation between the view-manifold and the circle-manifold. 
From Fig~\ref{normk_ratio}, KPLS-Norm Ratio supports the observation that we noted in the main paper, from Fig~\ref{regression_err}, that the lower layers in Model0 are more correlated to the circle-manifold than the higher layers. Except for Pool5, which encodes maximum correlation between the view-manifold and the circle-manifold. 


\medskip
\noindent{\emph 2) TPS-nonlinearity measure:} 
In this measure we learn a regularized Thin Plate Spline (TPS) non-linear mapping~\cite{duchon1977splines} between the unit circle manifold and each manifold $\mathcal{M}^k$. 
The mapping function ($\gamma$) can be written as
\[\gamma^k(\mathbf{x}) = \mathbf{C}^k  \cdot \psi(\mathbf{x}),\]
where $\mathbf{C}_{d \times (N+e+1)}$ is the mapping matrix, $e=2$, and 
the vector $\psi(x)=[\phi(|x-z_1|) \cdots \phi(|x-z_M|),1,x^T]^T$  represents a nonlinear kernel map from the  conceptual representation to a kernel induced space. The thin plate spline is defined as: $\phi(r) = r^3$ and $\{z_i\}_{i=1}^{M}$ are the set of center points. The solution for $\mathbf{C}^k$ can be obtained by directly solving the linear system:
\begin{equation}
\small
\begin{pmatrix} \mathbf{K}^l+\lambda \mathbf{I} & \mathbf{P}_x \\
 \mathbf{P}_t^T & \mathbf{0}_{(e+1) \times (e+1) } \end{pmatrix}_k  {\mathbf{C}^k}^T= \begin{pmatrix} \mathbf{A}_k \\ \mathbf{0}_{(e+1) \times d} \end{pmatrix},
\end{equation}
$\mathbf{A}$, $\mathbf{P}_x$ and $\mathbf{P}_t$ are defined for the $k-th$ set of object images as: $\mathbf{A}$ is a $N_k \times M$ matrix with $\mathbf{K}^l_{ij}=\phi(|x^k_i - z_j|), i=1,\cdots,N_k, j=1,\cdots,M, \mathbf{P}_x$ is a $N_k \times (e+1)$ matrix with $i$-th row $[1,\mathbf{x}^{k^T}_i]$, $\mathbf{P}_t$ is $M \times (e+1)$  matrix with $i$-th row $[1,\mathbf{z}^T_i]$. $A_k$ is a $N_k \times d$ matrix containing the set of images for manifold $\mathcal{M}^k$, \emph{i.e.} $\mathbf{A}_k=[\mathbf{y}_1^k,\cdots, \mathbf{y}^k_{N_k}]$. Solution for $\mathbf{C}^k$ is guaranteed under certain conditions on the basic functions used.

The reason for using TPS in particular is that the mapping has two parts, an affine part (linear polynomial) and a nonlinear part. Inquiring into the two parts gives an impression about the mapping, if it is mostly linear or nonlinear. We used the {\emph reciprocal-condition number (RCond)} of the submatrices of the coefficient matrix that correspond to the affine and the nonlinear part. 

While Fig~\ref{tps_poly} shows that the lower layers has more (better) conditioned linear mapping.  Fig~\ref{tps_cf} shows that the lower layer has complete stable mapping. This is expected since the lower layers have high dimensionality. At the same time, Fig~\ref{tps_cf} shows that the Convolution layers (Conv 3,4 and 5) have unstable nonlinear mappings. An additional observation is that fine-tuning against the pose labels increases the mapping stability (polynomial and non-polynomial). It is clear in Fig~\ref{tps_cf} that the $TPS-RCond(CF-nonPoly)$ has very small order of values ($10^{-11}$), therefore, we do not rely on it in our analysis.

\end{document}